\documentclass{article} 
\usepackage{iclr2027_preprint,times}

\usepackage{amsmath,amsfonts,bm}

\def\eqref#1{equation~\ref{#1}}

\def\1{\bm{1}}

\DeclareMathAlphabet{\mathsfit}{\encodingdefault}{\sfdefault}{m}{sl}
\SetMathAlphabet{\mathsfit}{bold}{\encodingdefault}{\sfdefault}{bx}{n}

\DeclareMathOperator*{\argmax}{arg\,max}

\usepackage{graphicx}
\usepackage{hyperref}
\usepackage{url}
\usepackage{booktabs}
\usepackage{multirow}
\usepackage{subcaption}
\usepackage{float}
\usepackage{placeins}

\title{Expression-Diverse References for Identity-Preserving Video Generation}

\author{Tianwen Fu \& Wenbin Teng \& Gonglin Chen \& Junyi Ouyang \& Haolin Xiong \& Yajie Zhao \\
Institute for Creative Technologies, University of Southern California \\
}

\definecolor{codexcommentcolor}{RGB}{170,0,140}
\newcounter{codexcomment}

\iclrpreprint
\begin{document}
\maketitle

\begin{abstract}

Identity-preserving video generation aims to maintain a subject's identity while synthesizing realistic videos. Yet a single reference portrait captures the subject's appearance under only one facial configuration. As expressions change, facial appearance can vary in highly identity-specific ways, leaving the subject's appearance under unseen expressions underdetermined by the reference alone. This expression-dependent variation also complicates evaluation: similarity to a neutral reference may decrease under strong expressions even for real images of the same person. We investigate this limitation from both generation and evaluation perspectives. First, we quantify how face-recognition similarity varies with expression intensity using controlled photographs and MEAD videos. We then construct a compact yet expressive reference gallery that captures diverse expression-dependent facial configurations. Matching against this gallery provides a more robust measure of identity similarity under expressive motion. To further expose performance degradation with expression intensity, we report identity similarity separately for mild, intense, and extreme expressions. For generation, we extend Stand-In to condition on our expression-diverse reference sets and develop a data-curation pipeline that extracts consistent yet diverse face crops from training videos. In practical settings where only a single portrait is available, we construct the reference set by synthesizing additional expressions with a pretrained facial reenactment model. On our controlled benchmark of 10 subjects and 48 prompts per subject, both real and synthesized reference sets outperform the evaluated baselines in identity similarity across all three expression-intensity regimes, with the largest improvements for extreme expressions. In this regime, nearest-gallery CurricularFace similarity increases from 60.95 for the strongest baseline to 74.80 with real reference sets and 68.36 with synthesized reference sets. These results show that representing an identity with an expression-diverse reference gallery substantially improves identity preservation under expressive motion.

\end{abstract}

\section{Introduction}

\begin{figure}[htbp]
    \centering
    \begin{subfigure}[t]{0.2\linewidth}
        \centering
        \includegraphics[width=\linewidth]{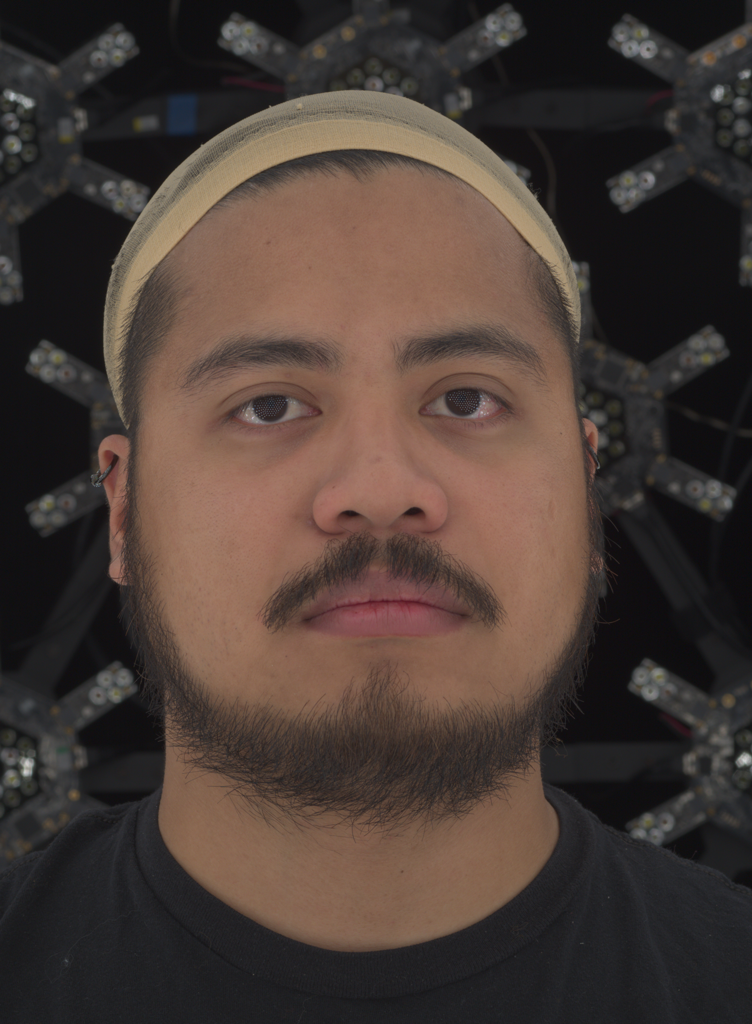}
        \caption*{\scriptsize Subject 1\\Neutral}
    \end{subfigure}\hfill
    \begin{subfigure}[t]{0.2\linewidth}
        \centering
        \includegraphics[width=\linewidth]{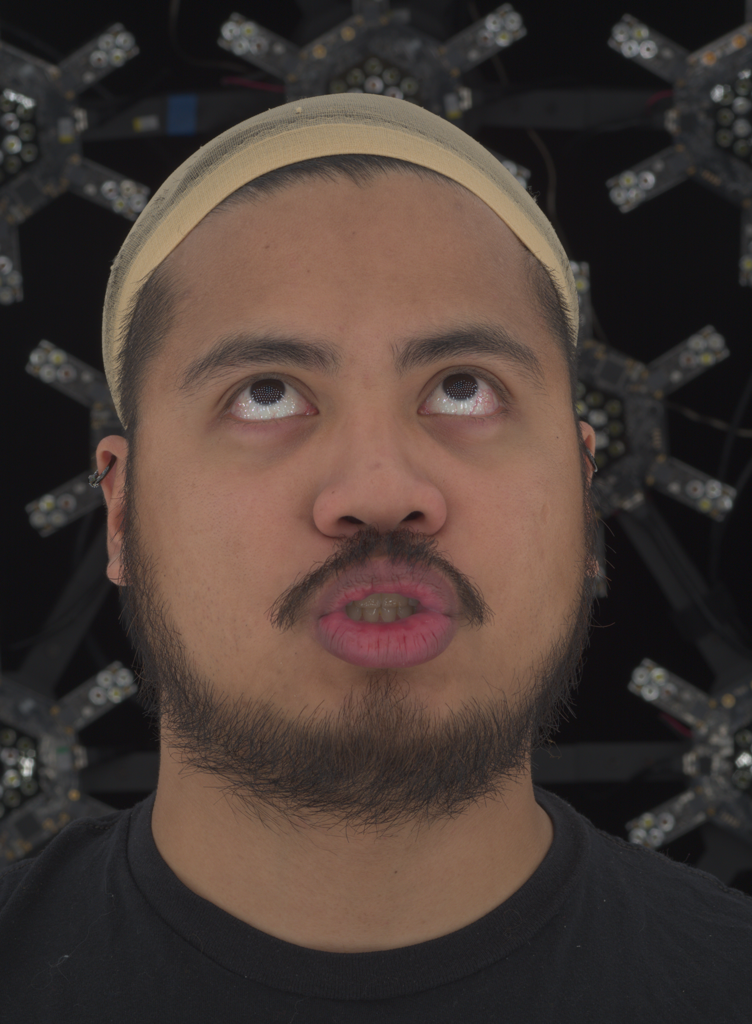}
        \caption*{\scriptsize Subject 1\\Expression 2}
    \end{subfigure}\hfill
    \begin{subfigure}[t]{0.2\linewidth}
        \centering
        \includegraphics[width=\linewidth]{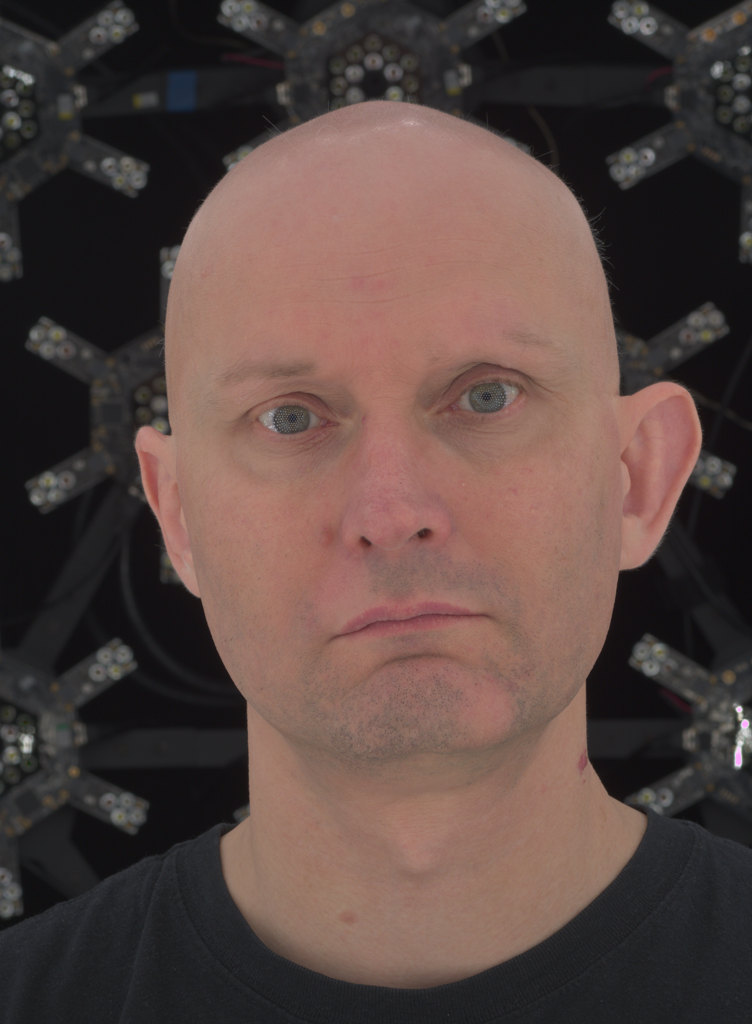}
        \caption*{\scriptsize Subject 2\\Neutral}
    \end{subfigure}\hfill
    \begin{subfigure}[t]{0.2\linewidth}
        \centering
        \includegraphics[width=\linewidth]{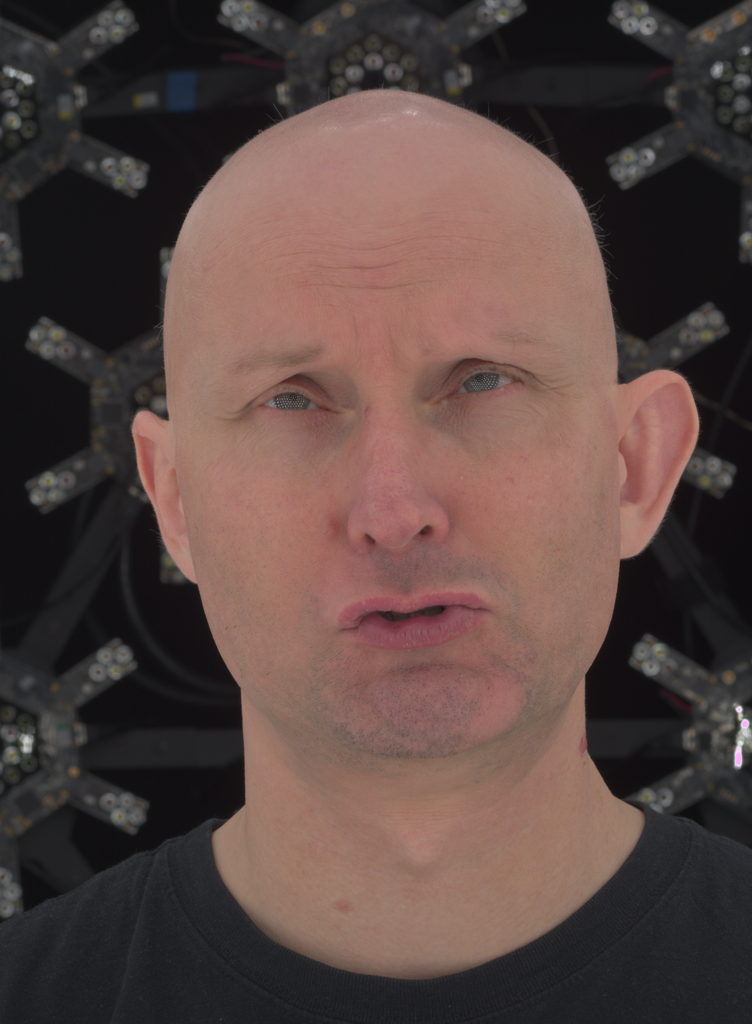}
        \caption*{\scriptsize Subject 2\\Expression 2}
    \end{subfigure}
    \caption{Expression-induced appearance changes are person-specific. Given the same expression instructions, the two subjects exhibit different changes in gaze, brow, and mouth geometry, illustrating information absent from a neutral reference.}
    \label{fig:personalized-expression-variation}
\end{figure}

Recent text-to-video models~\citep{wan2025,yang2024cogvideox,gao2025seedance} can synthesize realistic and diverse videos from natural-language prompts. Identity-preserving video generation extends this capability by introducing visual references to control the identity of the generated subject while following the text prompt. Recent methods build on pretrained video generators and incorporate reference-conditioning mechanisms to preserve the target identity throughout generation~\citep{Wang_2026_CVPR,xue2025stand}. Strong facial expressions pose a particular challenge: a person should remain recognizable even as their eyes, mouth, cheeks, and skin undergo substantial changes. A single portrait, as adopted by \citet{zhong2025concat,yuan2025identity}, captures only one facial configuration and provides limited evidence of these person-specific appearance changes (Figure~\ref{fig:personalized-expression-variation}).


This limitation affects both generation and evaluation. In generation, single-reference methods can lose identity fidelity as expression intensity (the degree to which a facial expression deviates from neutral) increases (Figure~\ref{fig:id-cur-nearest-vs-delta-psi}). In evaluation, comparing each generated face only with a neutral reference can conflate expression variation with identity error. We observe this effect even in real images: across 70 non-neutral photographs of 10 subjects, CurricularFace~\citep{huang2020curricularface} similarity to each subject's neutral reference decreases with expression intensity (Spearman $\rho=-0.701$) (Figure~\ref{fig:expression-identity-lightstage}). We observe a similar trend on a larger-scale video dataset, MEAD~\citep{kaisiyuan2020mead} (Table~\ref{tab:expression-identity-correlation}). Thus, single-reference similarity can be a suboptimal metric for identity consistency under strong expression changes.

We study expression-diverse references as a shared basis for conditioning and evaluation. For evaluation, we collect a standardized gallery of eight facial configurations per subject, comprising one neutral portrait and seven expressions (Figure~\ref{fig:lightstage-gallery}). We score each generated face by its maximum embedding cosine similarity to the subject's gallery, which provides a more expression-robust face similarity metric. We additionally report identity similarity separately for mild, intense, and extreme expressions, together with the fraction of generated frames in each regime. This makes failures at high expression intensity visible and helps interpret scores when methods produce different expression distributions.


For generation, we extend Stand-In~\citep{xue2025stand} to condition on multiple references of the same person. The model is an efficient 315M-parameter LoRA adapter with only 25k training steps at batch size 16. Expression-diverse references provide complementary observations of the subject's facial appearance that are absent from a single portrait. To train the model, we curate consistent and diverse face crops from OpenHumanVid~\citep{li2024openhumanvid} and vary the number of conditioning references during training. When multiple captures are unavailable, we use LivePortrait~\citep{guo2024liveportrait} to synthesize expressive references from a single neutral portrait, using driving expressions from a subject outside the evaluation set. This allows the same video generator to operate from a single portrait. Although identity similarity is lower than with real references, it still exceeds that of all evaluated baselines.


We evaluate on a controlled benchmark containing 10 subjects and 48 expression-focused prompts per subject. Both real and synthetic reference galleries achieve higher nearest-gallery identity similarity than all evaluated baselines in each of the three intensity regimes. Under extreme expressions, CurricularFace similarity, is 74.80 with real galleries and 68.36 with synthetic galleries, compared with 60.95 for the strongest baseline (Table~\ref{tab:curated-identity-results}). Ablations compare expression-diverse references with repeated neutral portraits and pose-diverse neutral references, examining the role of expression coverage beyond reference count (Table~\ref{tab:reference-diversity-ablation}). Across both recognizers and all evaluated baselines, our real-gallery model wins on at least 97.9\% of matched subject--prompt pairs. The win rate remains at least 96.1\% on the subset where our model produces greater peak expression intensity, providing additional evidence that identity gains can persist under stronger facial deformation (Table~\ref{tab:drive-matched-identity-wins}).


Our contributions are threefold:
\begin{itemize}
    \item We identify and quantify the dependence of single-reference face similarity on expression intensity in real imagery and introduce a standardized gallery-based multi-reference evaluation protocol stratified by expression intensity.
    \item We develop expression-diverse multi-reference conditioning and its data-curation pipeline, which enables the model to better capture person-specific facial deformations under expressive motion.
    \item We demonstrate improvements in identity similarity with both real expression reference galleries and reference galleries synthesized from a single portrait, both surpassing baselines in all expression regimes and especially the intense and extreme ones, even when multiple-expression captures are not available.
\end{itemize}

\section{Related Work}

\textbf{Video Generation Models} provide the generative backbone for human video synthesis. CogVideoX~\citep{yang2024cogvideox} combines a temporally compressive 3D VAE with an expert transformer to produce long, coherent text-conditioned clips. Wan2.2~\citep{wan2025wan22} introduces a two-expert architecture specialized for high- and low-noise denoising stages, providing a backbone for multiple downstream tasks. LTX-2~\citep{hacohen2026ltx} jointly models video and audio, while MiniMax H3~\citep{minimax2026h3} accepts multimodal character and motion references for video generation and editing. These general-purpose models provide strong visual priors, but text conditioning lacks precise control over the subject's identity in generated videos, whereas image-to-video generation generally requires a complete initial frame rather than just a reference portrait.



\textbf{Identity-Preserving Video Generation} conditions pretrained video generation backbones on reference images of one or more identities. The most common approach is to use a single reference image per identity and inject its features into the video generator. ConsisID~\citep{yuan2025identity} decomposes a reference face into low-frequency global features and high-frequency local details and injects them at different depths of a diffusion transformer. Concat-ID~\citep{zhong2025concat} instead concatenates VAE-encoded reference features with video latents, while Stand-In~\citep{xue2025stand} adds a lightweight conditional image branch with restricted self-attention and conditional position mapping. Phantom~\citep{phantom2025} learns joint text--image injection from text--image--video triplets.

\textbf{Benchmarking Video Generation Models} for facial identity similarity typically relies on a single reference. VBench~\citep{huang2024vbench} evaluates video generation along 16 disentangled dimensions. Its subject-consistency metric compares DINO features across frames of a generated video, measuring temporal stability but not facial identity relative to a reference person. OpenS2V-Eval~\citep{yuan2025opens2v} more directly evaluates subject-to-video generation. For human identity, its FaceSim protocol detects faces and computes CurricularFace~\citep{huang2020curricularface} cosine similarity to the input face. However, as we show below, this single-reference comparison can penalize genuine identity-preserving expression changes when the generated face departs from the reference expression. Our benchmark instead compares each frame with the nearest member of a standardized expression gallery and reports results stratified by expression intensity.


\textbf{Facial Reenactment and Related Tasks.} Facial reenactment animates a source portrait using head pose and expressions extracted from a driving video. LivePortrait~\citep{guo2024liveportrait} extracts appearance features from a single source image, estimates implicit 3D keypoints for the source and driving frames, warps the source features according to their motion, and decodes the result into animated frames. Person-specific 3D approaches such as GaussianAvatars~\citep{qian2024gaussianavatars} instead rig optimized 3D Gaussians to a parametric face model, and video face-swapping systems such as OmniFace~\citep{guo2026omniface} transfer a source identity onto an existing target video while preserving its motion and scene. These tasks receive an explicit driving signal and are distinct from our text-driven generation setting.

\section{Evaluating Identity Fidelity at High Expression Intensity}
\begin{figure}[htbp]
    \centering
    \begin{tabular}{@{}*{7}{c@{\hspace{0.005\linewidth}}}c@{}}
        \includegraphics[width=0.12\linewidth]{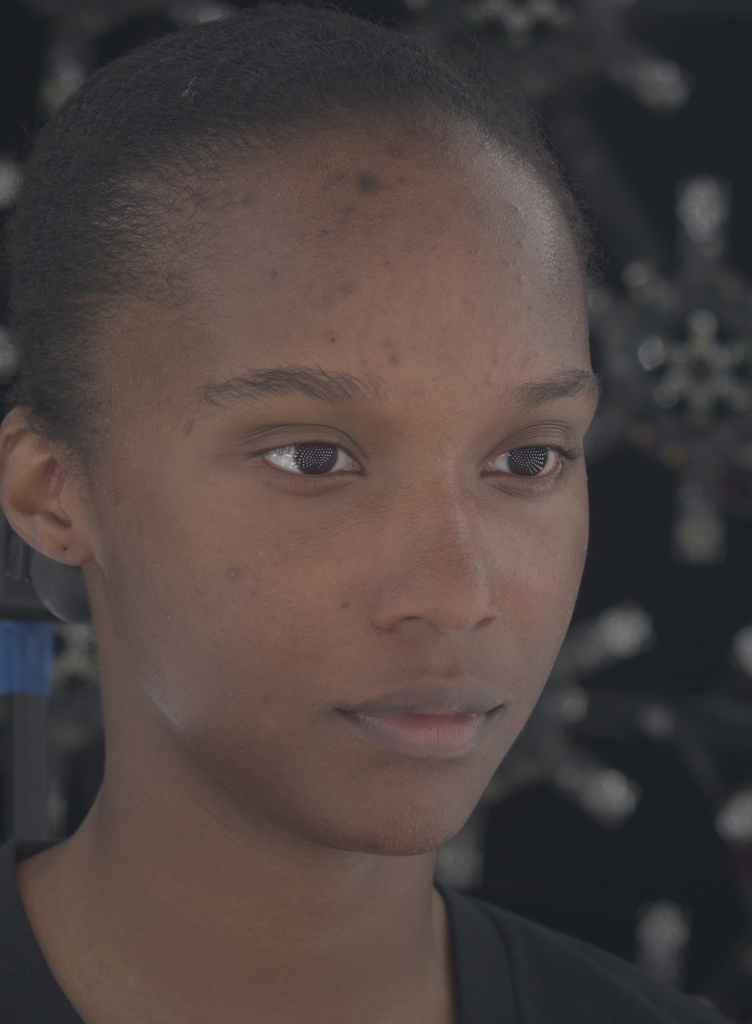} &
        \includegraphics[width=0.12\linewidth]{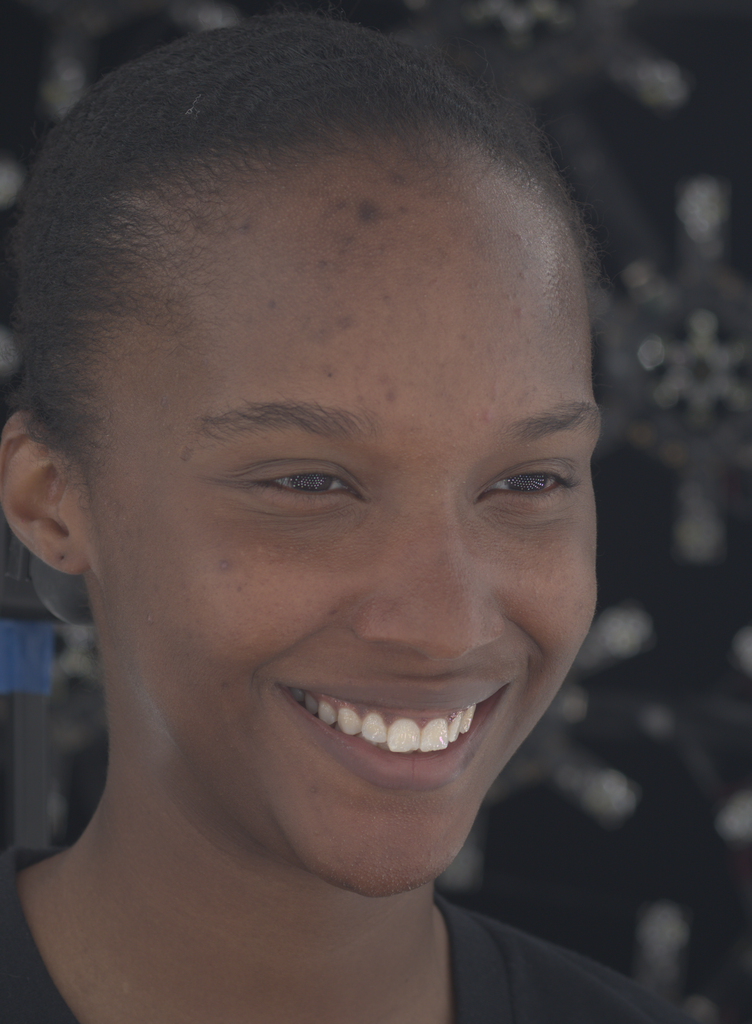} &
        \includegraphics[width=0.12\linewidth]{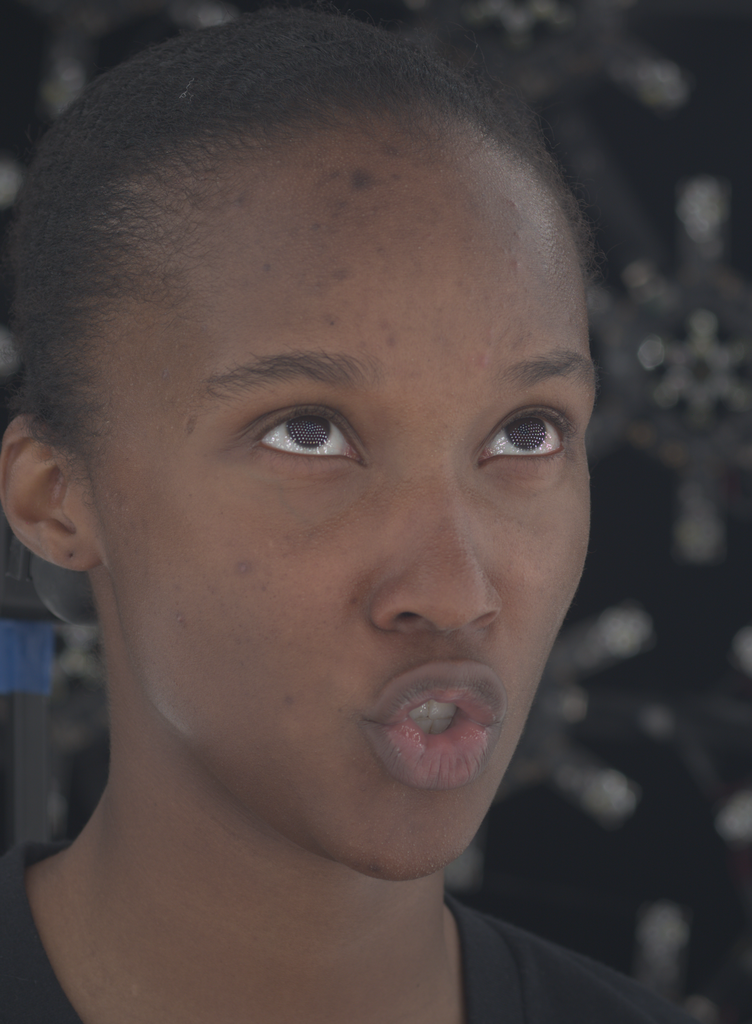} &
        \includegraphics[width=0.12\linewidth]{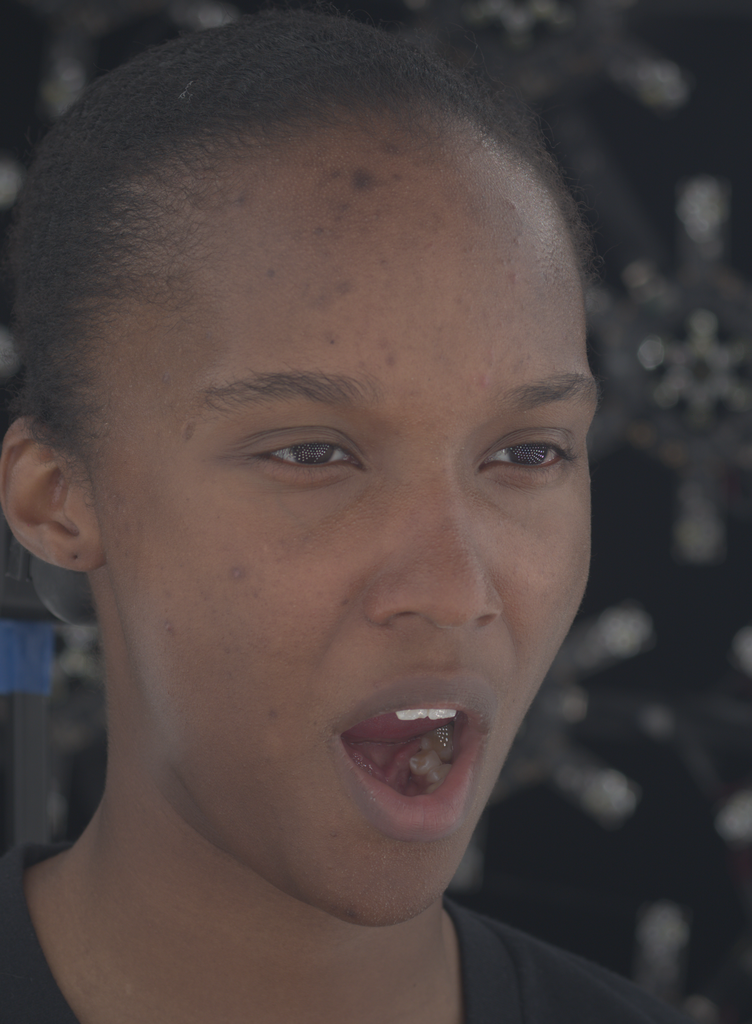} &
        \includegraphics[width=0.12\linewidth]{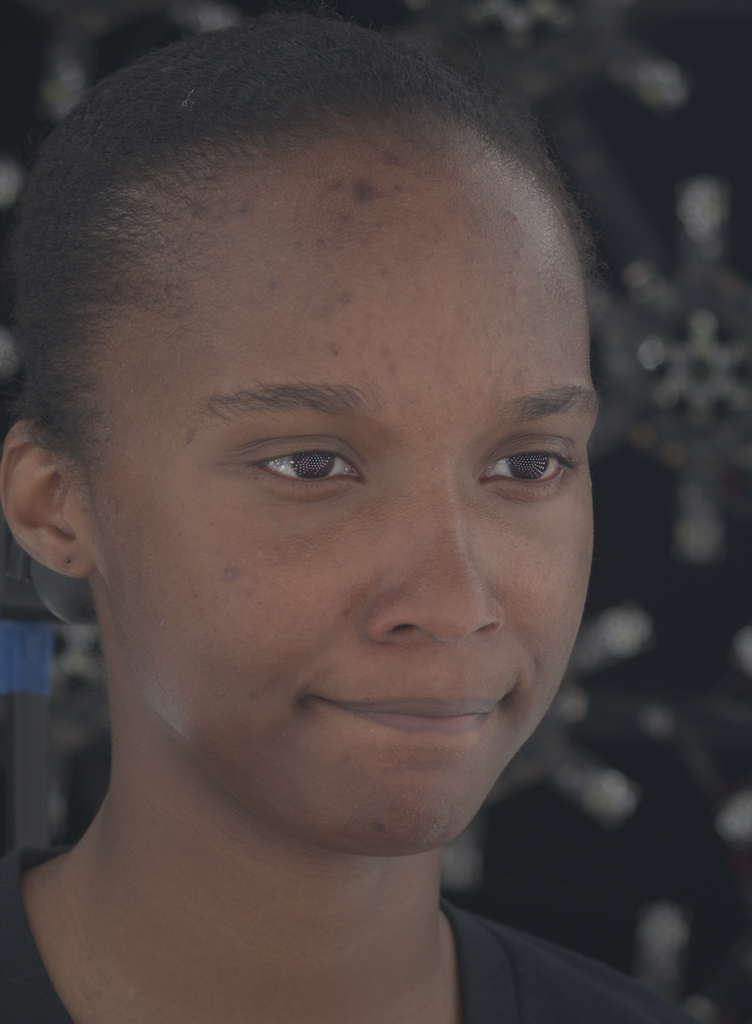} &
        \includegraphics[width=0.12\linewidth]{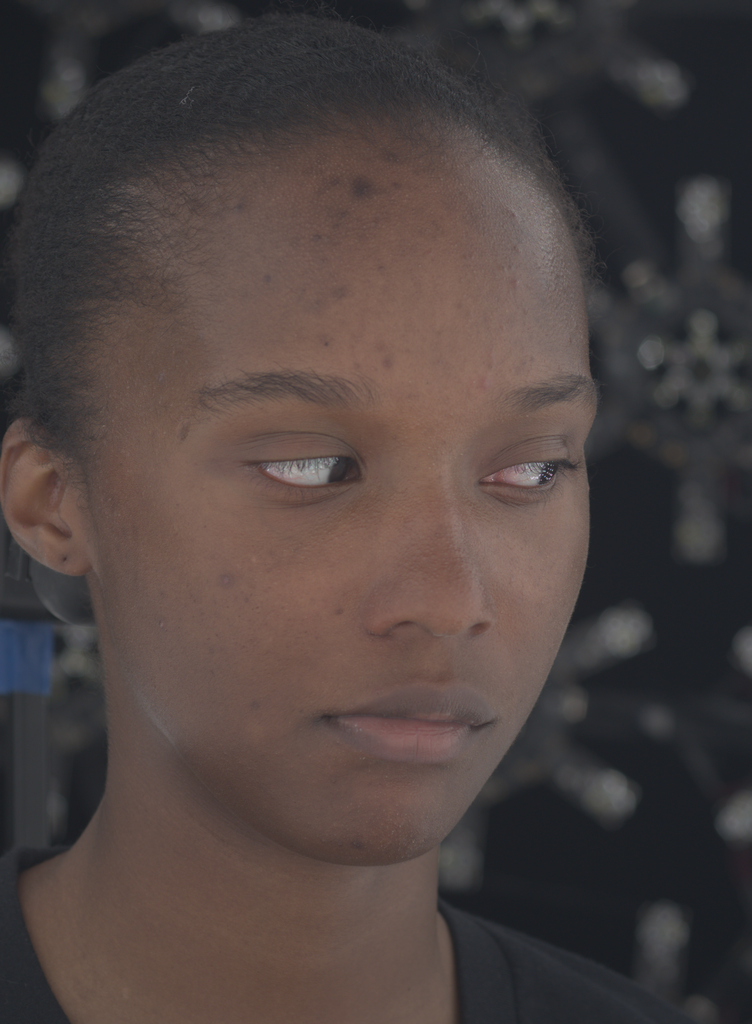} &
        \includegraphics[width=0.12\linewidth]{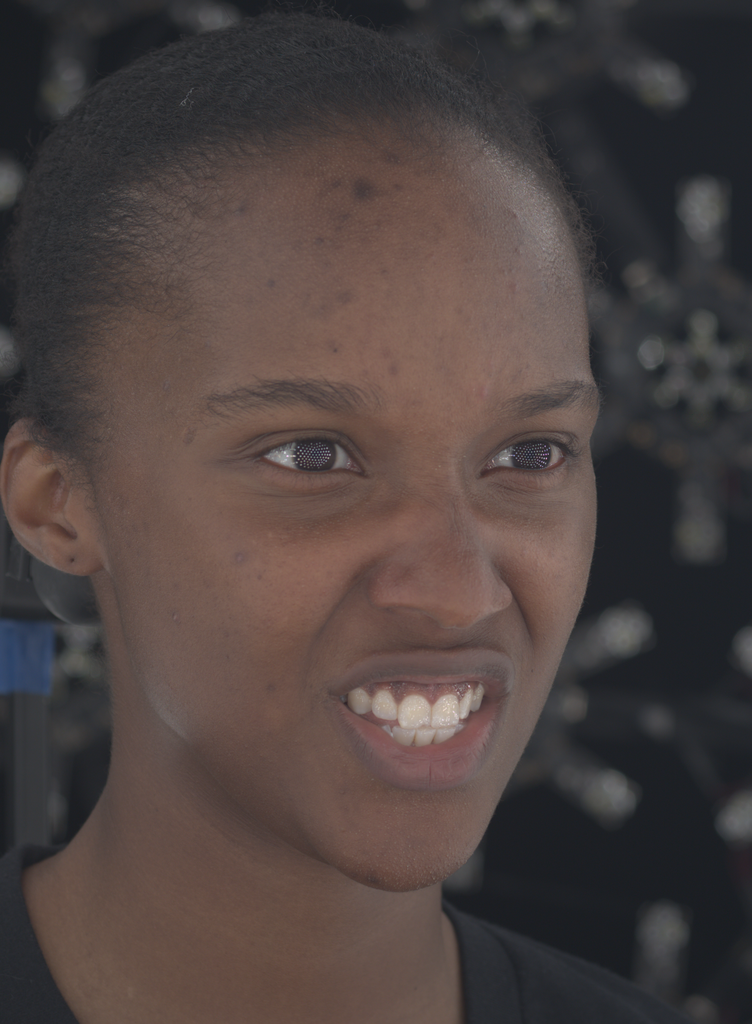} &
        \includegraphics[width=0.12\linewidth]{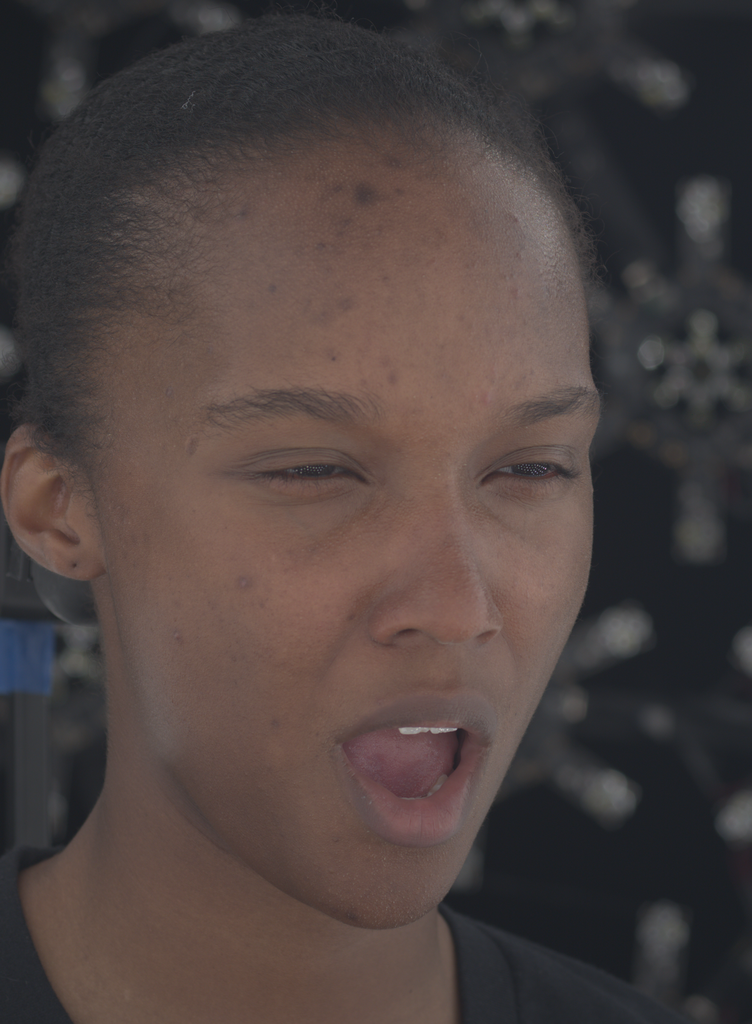} \\
        \scriptsize Neutral (0) &
        \scriptsize Expression 1 & 
        \scriptsize Expression 2 & 
        \scriptsize Expression 3 & 
        \scriptsize Expression 4 & 
        \scriptsize Expression 5 & 
        \scriptsize Expression 6 & 
        \scriptsize Expression 7 
    \end{tabular}
    \caption{Example gallery for one of the 10 subjects with 8 expressions including neutral. All subjects were asked to perform the same set of expressions, whose instructions are listed in Appendix~\ref{sec:gallery-expression-instructions}. The gallery captures diverse expression-dependent facial configurations to support more robust identity evaluation and provide richer information for video generation models.}
    \label{fig:lightstage-gallery}
\end{figure}

\begin{figure}[htb]
    \centering
    \begin{subfigure}[t]{0.4\linewidth}
        \centering
        \includegraphics[width=\linewidth]{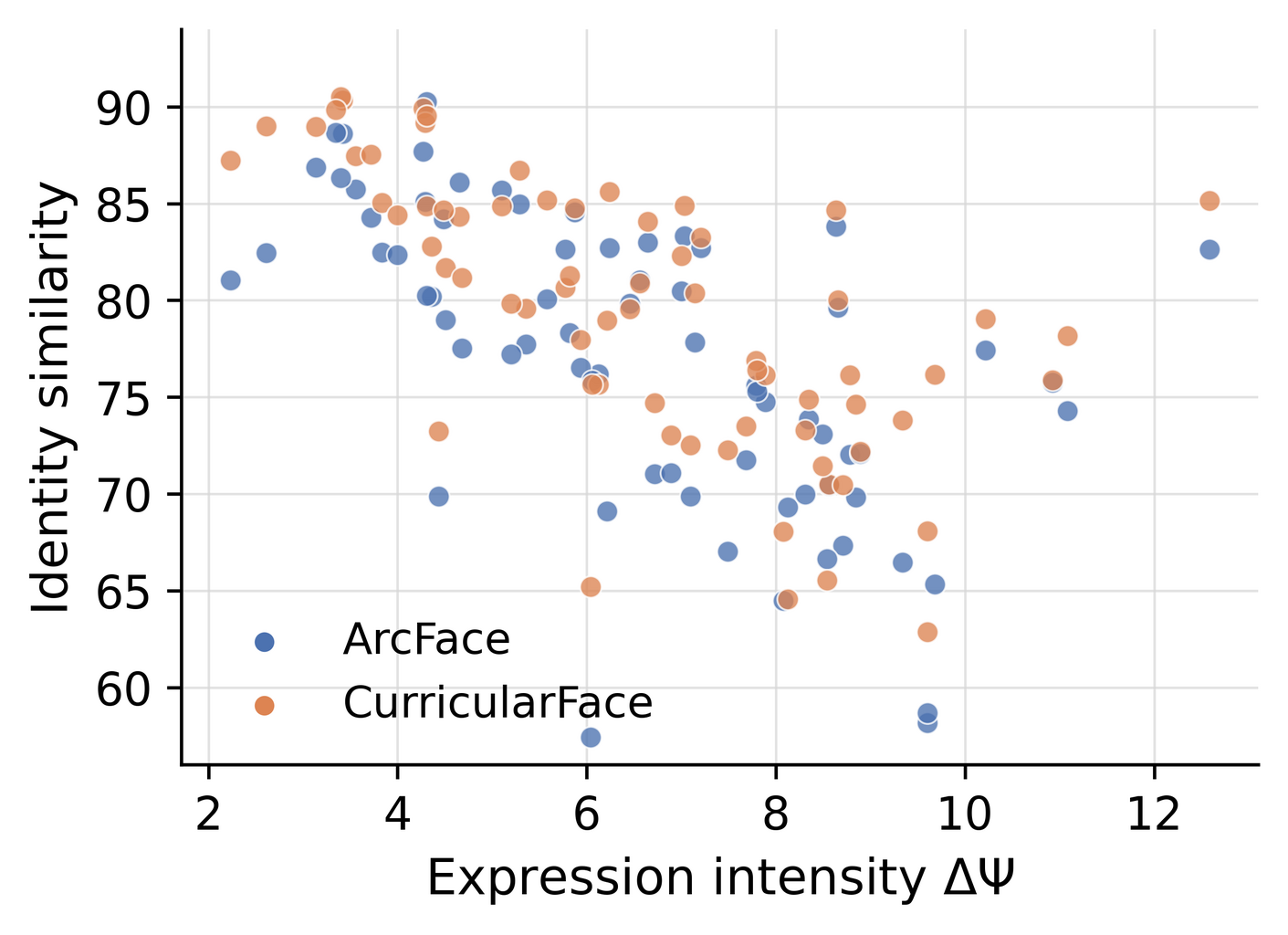}
        \caption{Lightstage}
        \label{fig:expression-identity-lightstage}
    \end{subfigure}\hspace{0.1\linewidth}
    \begin{subfigure}[t]{0.4\linewidth}
        \centering
        \includegraphics[width=\linewidth]{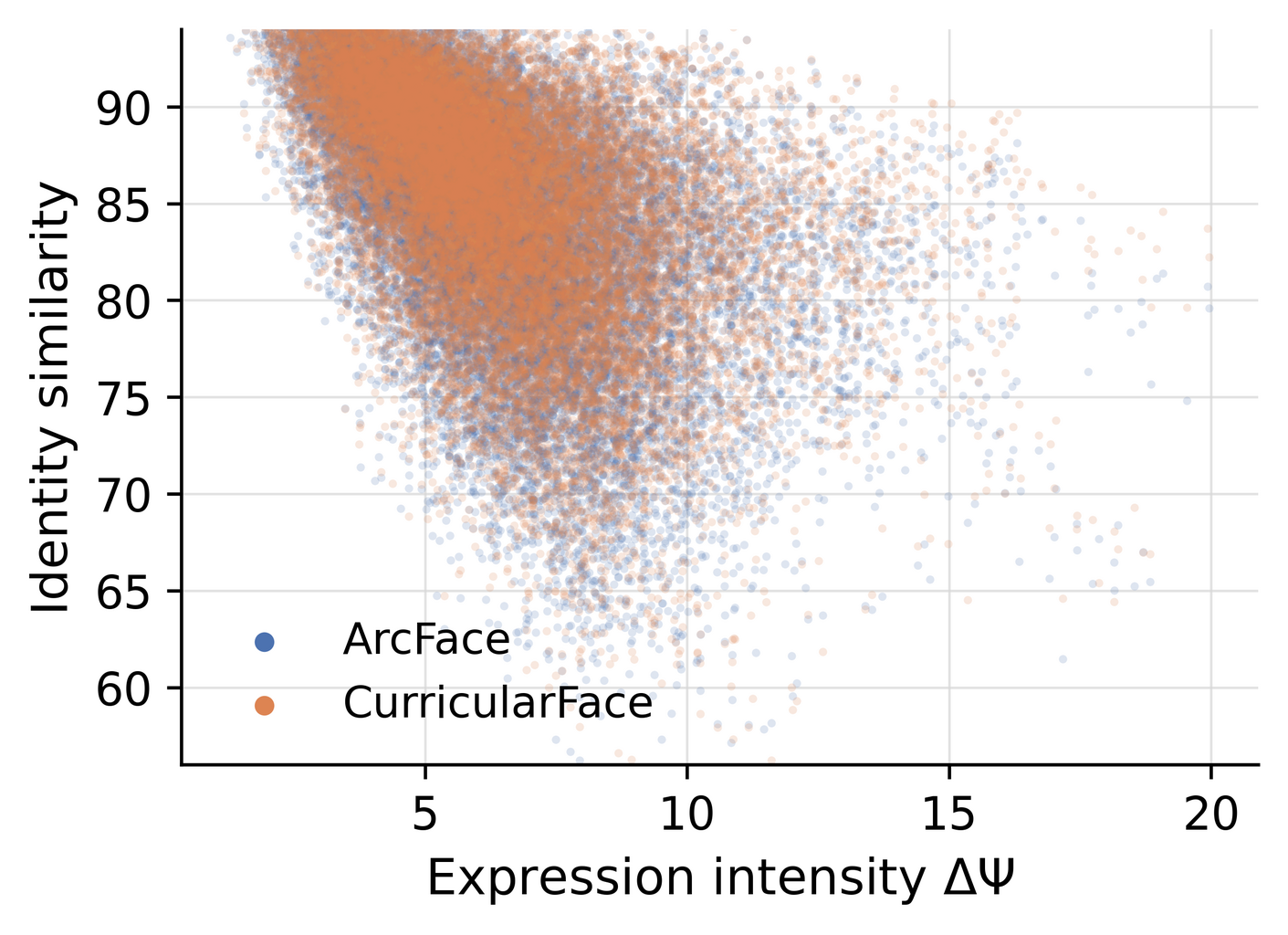} 
        \caption{MEAD}
        \label{fig:expression-identity-mead}
    \end{subfigure}
    \caption{ArcFace and CurricularFace similarity versus expression intensity on real images. Each point represents a non-neutral photograph (Lightstage) or video frame (MEAD), measured relative to the subject's neutral reference. MEAD preprocessing is described in Appendix~\ref{sec:mead-processing}.}
    \label{fig:expression-identity}
\end{figure}

Previous work evaluates facial identity by comparing a generated face with a single reference image of the same subject in the embedding space of a pretrained face-recognition model. However, we show that the face similarity score is entangled with the expression variance. We collect a gallery of reference images $\mathcal{G}=\{I_{i, e}\}$ for 10 subjects $i\in \{1, \ldots, 10\}$ and 8 expressions $e\in \{0, \ldots, 7\}$, where $e=0$ is the neutral expression. The images were taken in a light stage with a fixed frontal camera position and flat lighting setup, as shown in Figure~\ref{fig:lightstage-gallery}. For each subject $i$, we compare each non-neutral image ($e\neq 0$) with the neutral image ($e=0$), computing cosine similarity separately with ArcFace~\citep{deng2019arcface} ($s_{Arc}$) and CurricularFace~\citep{huang2020curricularface} ($s_{Cur}$). We measure expression intensity as $\Delta\Psi_{i, e} = \|\Psi_{i, e} - \Psi_{i, 0}\|_2$, where $\Psi_{i, e}$ is estimated using SMIRK~\citep{SMIRK:CVPR:2024} (Appendix~\ref{sec:face-pipeline}). As shown in Figure~\ref{fig:expression-identity}, both identity metrics decrease as expression intensity increases, both in our collected dataset and in the publicly available MEAD dataset~\citep{kaisiyuan2020mead}. To further quantify this, we report both Pearson's $r$ and Spearman's $\rho$ in Table~\ref{tab:expression-identity-correlation}.

\begin{table}[htbp]
    \centering
    \caption{Correlation between expression intensity $\Delta\Psi$ and identity similarity on both datasets.}
    \label{tab:expression-identity-correlation}
    \resizebox{0.8\linewidth}{!}{%
    \begin{tabular}{lrrrrr}
        \toprule
        & & \multicolumn{2}{c}{ArcFace} & \multicolumn{2}{c}{CurricularFace} \\
        \cmidrule(lr){3-4}\cmidrule(lr){5-6}
        Dataset & $n$ & Pearson's $r$ & Spearman's $\rho$ & Pearson's $r$ & Spearman's $\rho$ \\
        \midrule
        Lightstage & 70 & $-0.587$ & $-0.662$ & $-0.636$ & $-0.701$ \\
        MEAD & 24,190 & $-0.521$ & $-0.590$ & $-0.496$ & $-0.567$ \\
        \bottomrule
    \end{tabular}%
    }
\end{table}

Therefore, for recognizer $R \in \{\mathrm{Arc},\mathrm{Cur}\}$, we define the nearest-gallery identity similarity as $s_R^*(\hat{I}_i) = \max_{e\in\{0,\ldots,7\}} s_R(I_{i,e}, \hat{I}_i)$\footnote{We report similarities on a 100-point scale, following the convention of \citet{Wang_2026_CVPR}.}, where $\hat{I}_i$ is a frame generated for subject $i$. This provides a more robust measure of identity fidelity when the expression intensity is high.

\begin{figure}[htb]
    \centering
    \includegraphics[width=0.5\linewidth]{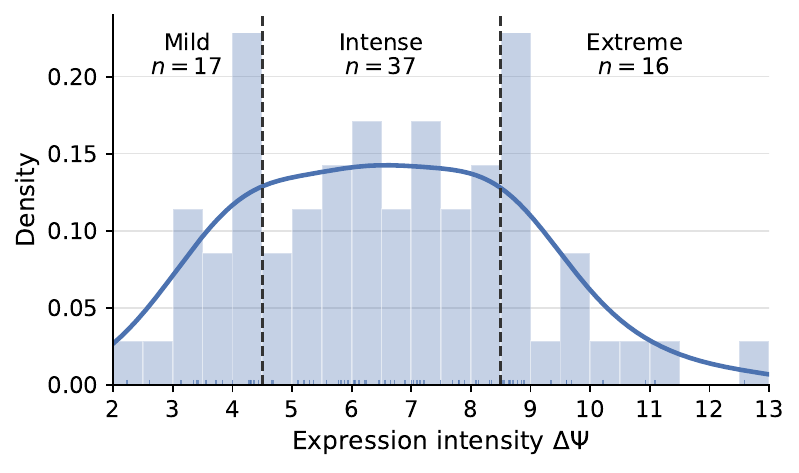}
    \caption{Distribution of expression intensity $\Delta\Psi$ across the 70 non-neutral light-stage photographs. The solid curve is a kernel density estimate of the density histogram, and marks along the horizontal axis denote individual photographs.}
    \label{fig:expression-intensity-distribution}
    \vspace{-1em}
\end{figure}

Furthermore, we report the nearest-gallery identity similarities for three ranges of $\Delta\Psi$: mild ($\Delta\Psi < 4.5$), intense ($4.5 \leq \Delta\Psi < 8.5$), and extreme ($\Delta\Psi \geq 8.5$). As shown in Figure~\ref{fig:expression-intensity-distribution}, the thresholds approximately correspond to the first and third quartiles of $\Delta\Psi$ across the 70 non-neutral real photographs, which already depict deliberately intense expressions. This places the majority of photographs in the intense regime while retaining similarly sized mild and extreme tails.

We combine the six basic emotions proposed by \citet{Ekman1992Facial} with four environments and two triggers to create 48 prompts per subject (Appendix~\ref{sec:eval-prompts}). The prompted expressions differ from the reference configurations to discourage direct copying.

\section{Training Multi-Reference Subject-to-Video Models}
\label{sec:method}

\textbf{Baselines.} We provide the neutral portrait to the single-reference subject-to-video baselines: ConsisID~\citep{yuan2025identity}, Concat-ID~\citep{zhong2025concat}, Stand-In~\citep{xue2025stand}, and Phantom~\citep{phantom2025}. For Gen-4.5~\citep{runway2025gen45}, we use image-to-video mode through the official API, with the text prompt and the neutral portrait as the initial frame. For AnyID~\citep{Wang_2026_CVPR}, which accepts up to five references, we provide the neutral portrait and four other expression images (1, 3, 5, 7). We use official checkpoints for the locally run baselines. For Stand-In, we use its adapter for Wan 2.2.

\begin{figure}[htb]
    \centering
    \begin{subfigure}[t]{0.245\linewidth}
        \centering
        \begin{tabular}{@{}c@{\hspace{0.006\textwidth}}c@{}}
            \includegraphics[height=1.30cm]{fig/lightstage_gallery/JadeMcBride_01.png} &
            \includegraphics[height=1.30cm]{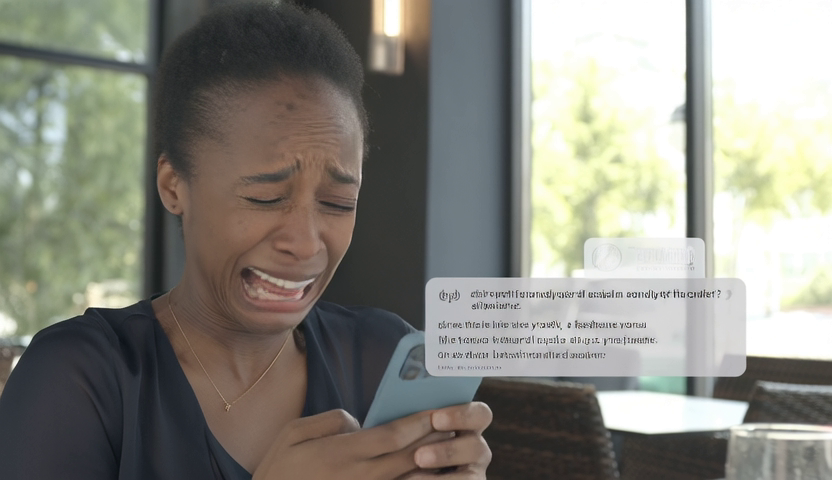} \\
            \scriptsize Reference & \scriptsize Stand-In
        \end{tabular}
        \caption{Distorted mouth}
        \label{fig:failure-broken-mouth}
    \end{subfigure}\hfill%
    \begin{subfigure}[t]{0.245\linewidth}
        \centering
        \begin{tabular}{@{}c@{\hspace{0.006\textwidth}}c@{}}
            \includegraphics[height=1.30cm]{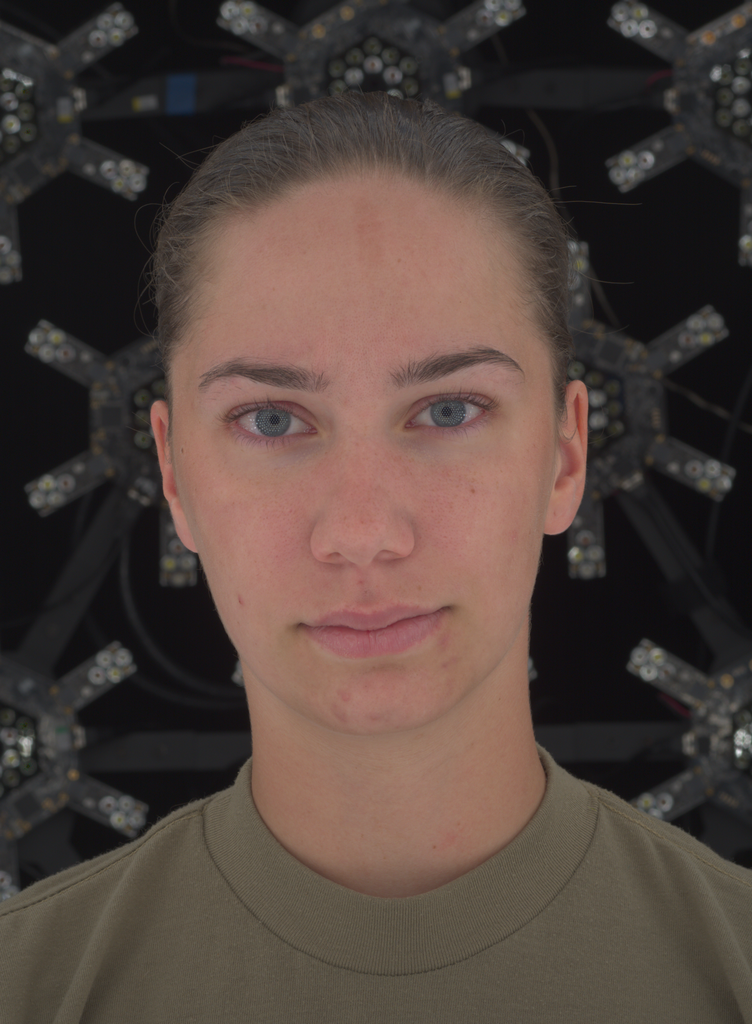} &
            \includegraphics[height=1.30cm]{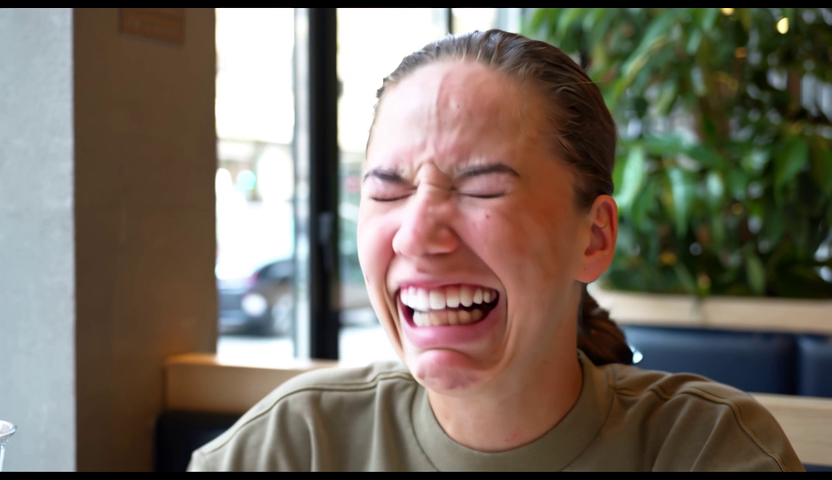} \\
            \scriptsize Reference & \scriptsize Phantom
        \end{tabular}
        \caption{Distorted glabella}
        \label{fig:failure-broken-glabella}
    \end{subfigure}\hfill%
    \begin{subfigure}[t]{0.50\linewidth}
        \centering
        \begin{tabular}{@{}c@{\hspace{0.006\textwidth}}c@{\hspace{0.025\textwidth}}c@{\hspace{0.006\textwidth}}c@{}}
            \includegraphics[height=1.30cm]{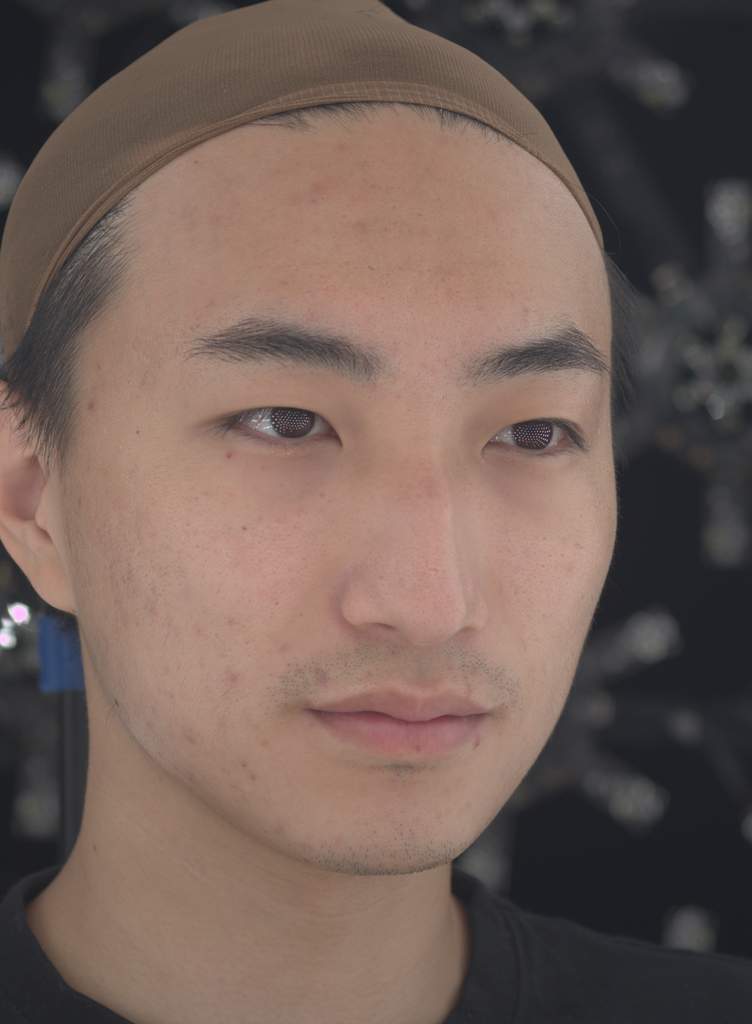} &
            \includegraphics[height=1.30cm]{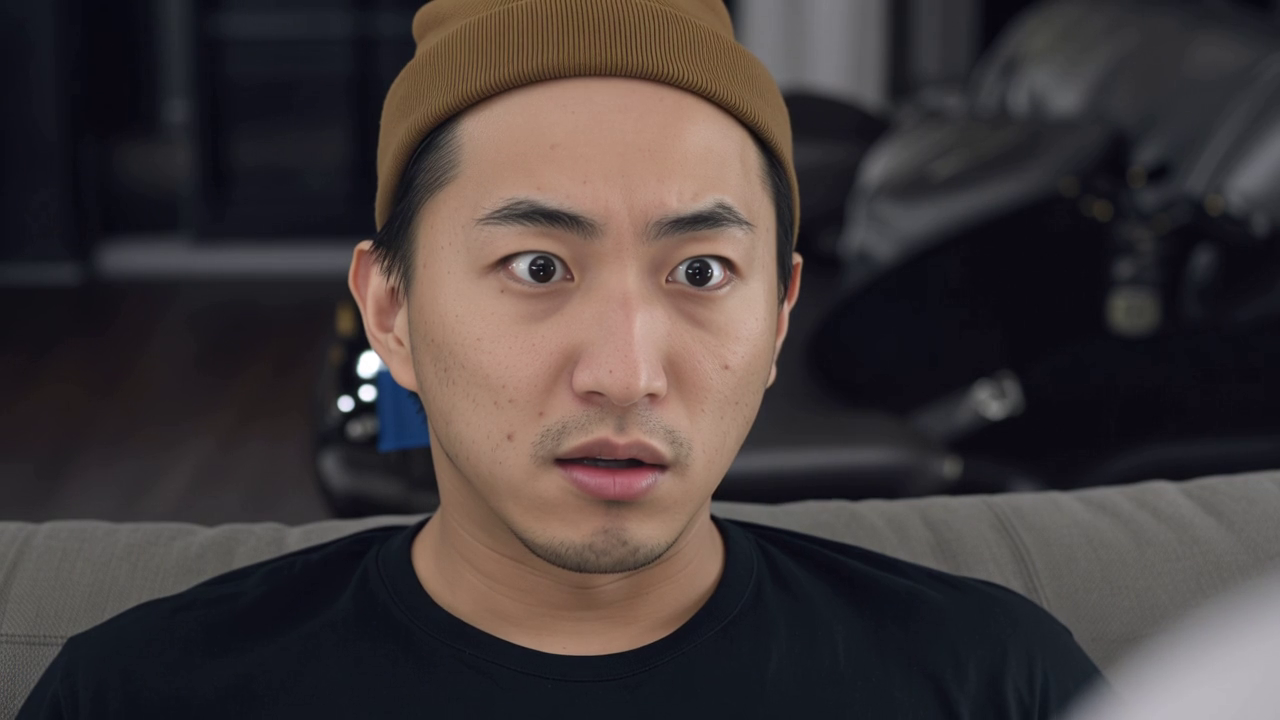} &
            \includegraphics[height=1.30cm]{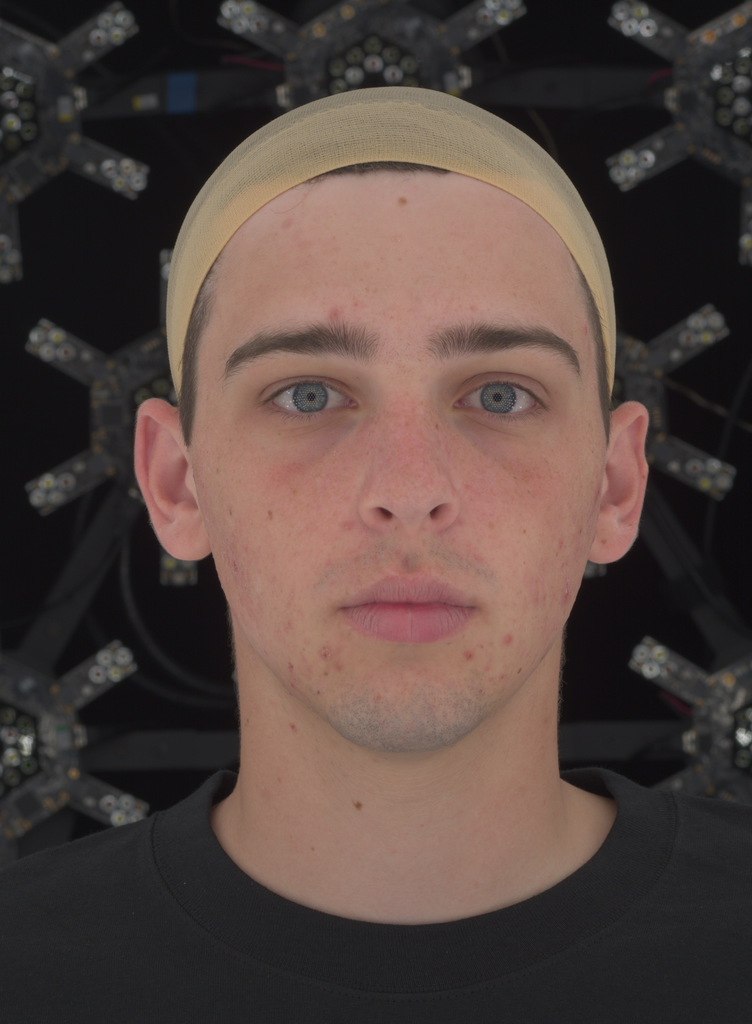} &
            \includegraphics[height=1.30cm]{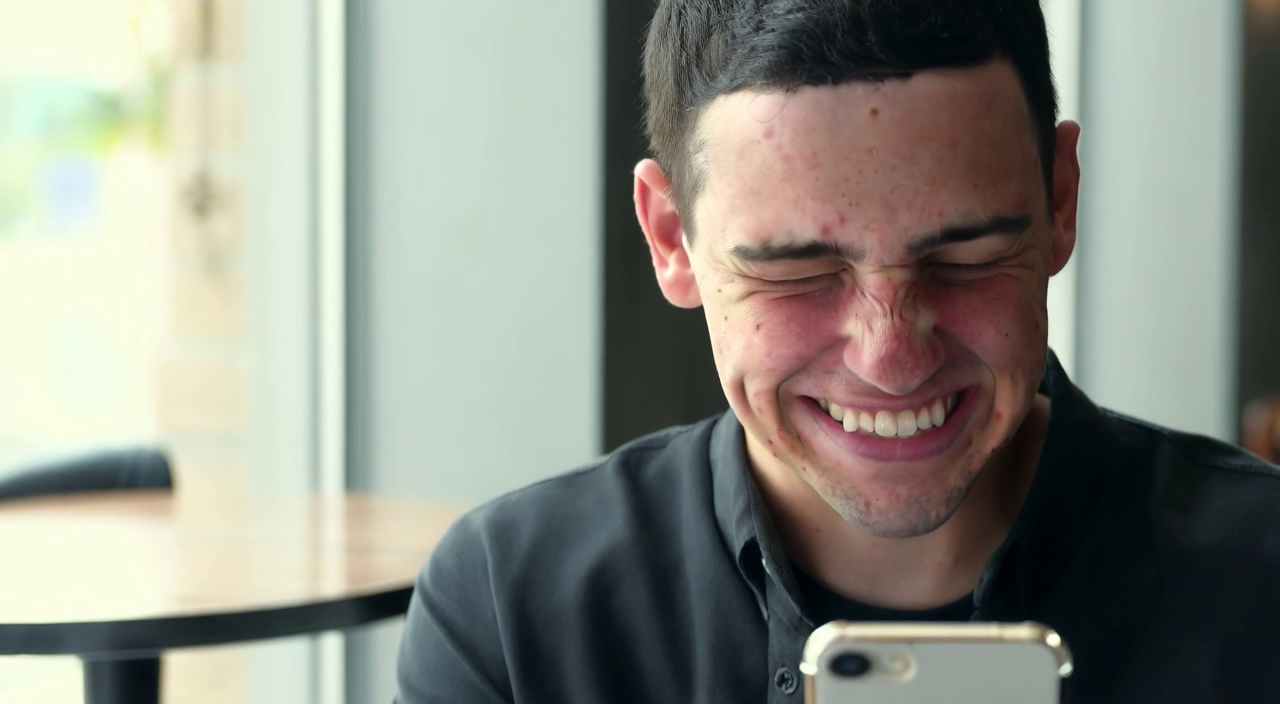} \\
            \scriptsize Reference & \scriptsize Gen-4.5 &
            \scriptsize Reference & \scriptsize AnyID
        \end{tabular}
        \caption{Incorrect facial features}
        \label{fig:failure-incorrect-features}
    \end{subfigure}
    \caption{Representative facial artifacts produced by existing identity-preserving video generation methods under expressive motion. Each generated frame is shown beside the neutral reference portrait provided to the method.}
    \label{fig:failure-examples}
\end{figure}

\begin{figure}[htb]
    \centering
    \includegraphics[width=0.8\linewidth]{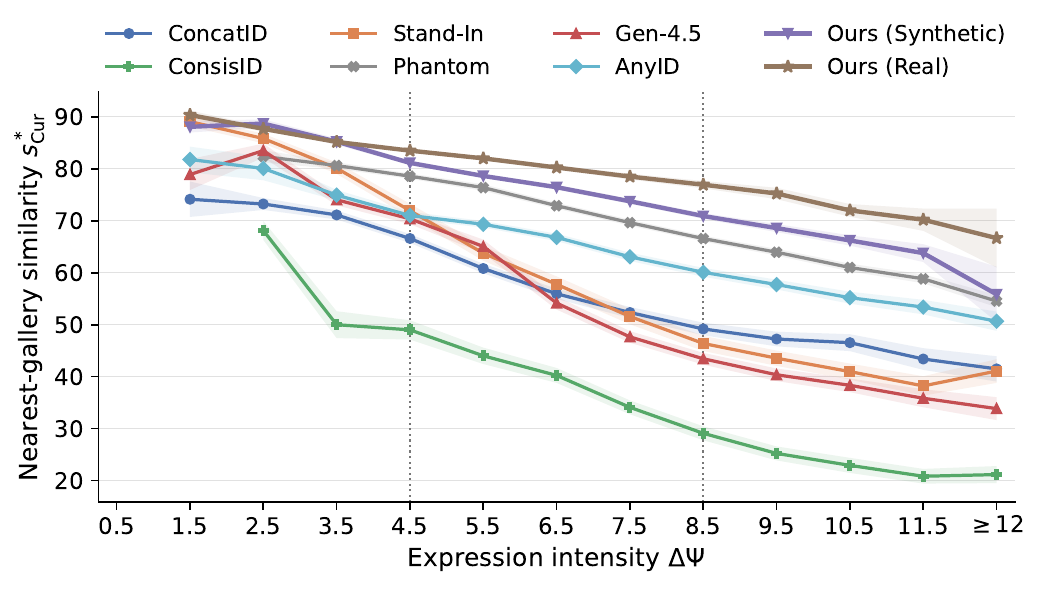}
    \caption{Distribution of nearest-gallery CurricularFace identity similarity in different bins of expression intensity $\Delta\Psi$. We use all generated frames from the baselines and our method, run SMIRK~\citep{SMIRK:CVPR:2024} to obtain the expression vector $\Psi$, and measure $\Delta\Psi = \|\Psi - \Psi_{\text{neutral}}\|_2$ against the neutral reference (see Appendix~\ref{sec:face-pipeline} for details). Each point gives the mean identity similarity within a unit-width $\Delta\Psi$ bin, connected across bins; the final point aggregates the open-ended $\Delta\Psi \geq 12$ bucket. Shaded regions denote 95\% confidence intervals computed using clip-clustered standard errors. Dotted lines mark the thresholds between mild, intense, and extreme expressions.}
    \label{fig:id-cur-nearest-vs-delta-psi}
\end{figure}

\textbf{Failure Mode of Existing Methods.} We analyze identity preservation across all generated frames at different expression intensities.  Figure~\ref{fig:id-cur-nearest-vs-delta-psi} shows that the single-reference models exhibit a substantial drop in identity similarity when the expression intensity $\Delta\Psi$ is large despite high similarity at low $\Delta\Psi$. Some common artifacts for these models are presented in Figure~\ref{fig:failure-examples}. The multi-reference model AnyID~\citep{Wang_2026_CVPR} shows a more stable curve, benefiting from our expression-diverse references, but still has low identity similarity across all intensity regimes.

\textbf{Data Curation.} We use the OpenHumanVid~\citep{li2024openhumanvid} dataset for training our multi-reference subject-to-video models. For each video, we sample 10 evenly spaced frames and use SCRFD~\citep{guo2021sample} to detect the face. We retain the largest detected face with a detection score above $0.5$ and crop it using a bounding box expanded by a factor of $1.6$. We compute ArcFace embeddings for the retained crops and cluster them by linking together any two crops with cosine similarity above $0.35$. Videos with more than 2 clusters or more than 40\% of the crops outside the dominant cluster are treated as multi-subject videos and are discarded, and we keep the largest cluster for the remaining videos. If more than eight crops remain, we select a diverse subset of eight. Starting from the highest-quality crop $i_0=\argmax_{j}Q(I_j)$, we greedily add the farthest remaining crop $i_{k+1}$ until eight crops have been selected: \[
i_{k+1} = \argmax_{j} \min_{j' \in \{0, 1, \ldots, k\}} 0.75\cdot(1-s_{Arc}(I_j, I_{i_{j'}})) + 0.25\cdot\Delta t(j, i_{j'})
\] where $Q(I_j)$ is the quality score, $s_{Arc}(I_j, I_{i_{j'}})$ is the ArcFace cosine similarity, and $\Delta t(j, i_{j'})$ is the temporal distance between the two frames, normalized to $[0, 1]$.

\begin{figure}[t]
    \centering
    \includegraphics[width=\textwidth]{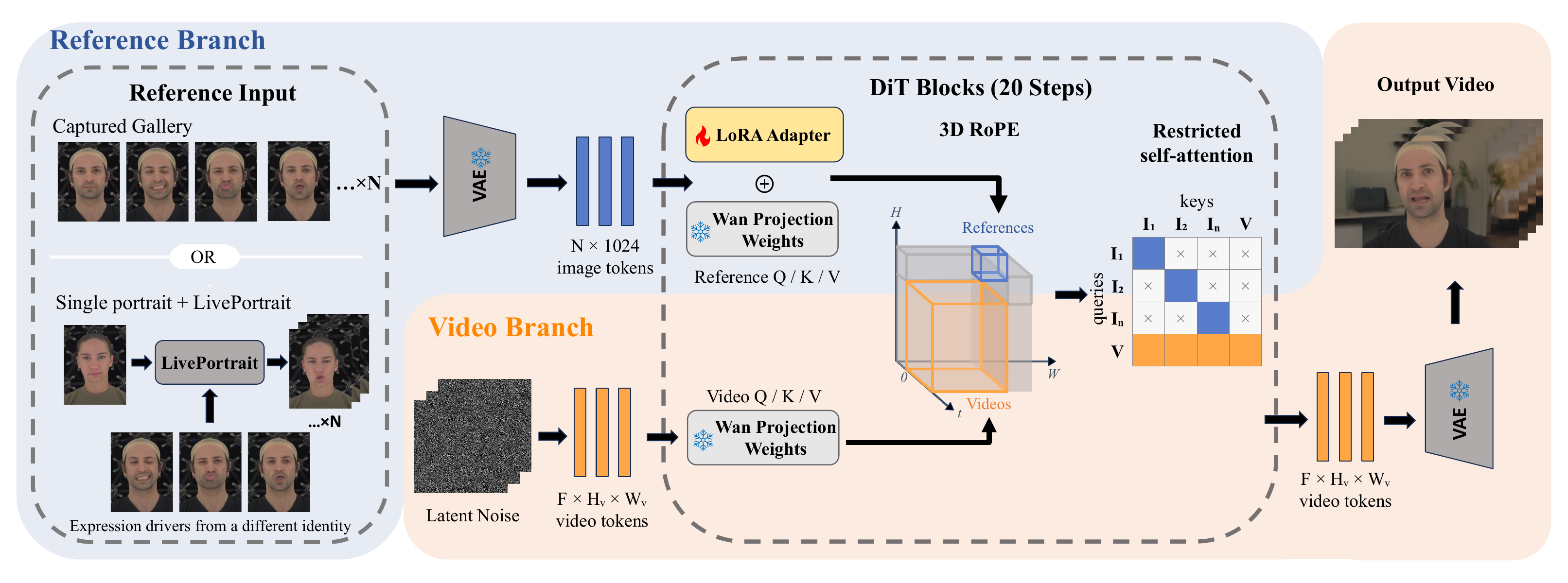}
    \caption{Overview of our multi-reference subject-to-video generation pipeline.}
    \label{fig:pipeline}
\end{figure}

\textbf{Architecture and Training.} The pipeline for our method is shown in Figure~\ref{fig:pipeline}. We build on Stand-In~\citep{xue2025stand}, a single-reference identity-preserving adapter for Wan2.2-T2V-A14B~\citep{wan2025}. It adds rank-128 LoRA adapters to the query, key, and value projections of each self-attention block, applied only to reference-image tokens, while the backbone parameters, including those used to process video tokens, remain frozen. We extend this architecture to accept multiple reference images. We independently resize and VAE-encode each of the $N$ references at $512 \times 512$ resolution, yielding $1024$ tokens per reference. All references share diffusion timestep $0$ and the same grid of 3D RoPE~\citep{su2021roformer} coordinates at temporal index $-1$ and spatial indices starting beyond the video grid, namely $\{-1\} \times [H_V, H_V + H_I) \times [W_V, W_V + W_I)$, whereas the video tokens are placed on $[0, F) \times [0, H_V) \times [0, W_V)$, where $F=21$ is the number of latent frames, compressed by the VAE from $81$ frames. Reference tokens attend only within their own image, while video tokens attend to all tokens. The attention mask over the concatenated tokens is \[
M_{Q, K}(i, j) = \begin{cases}
    1 & \text{if } i \text{ is a video token} \\
    1 & \text{if } i, j \text{ are image tokens from the same reference image} \\
    0 & \text{otherwise.}
\end{cases}
\]
The reference tokens do not attend to the text prompt in the cross-attention layers, and are discarded before untokenization and VAE decoding. We train with 400k randomly selected unique clips. During training, we shuffle the references. At each step, with probability $0.5$, we use all $N$ available images; otherwise, we draw $k$ uniformly from $\{1,\ldots,N\}$ and randomly select a subset of size $k$. This improves robustness to different reference counts. We use an effective batch size of 16.

\textbf{Inference Pipeline with a Single Reference.} Our model provides the best results when multiple references are provided, but when only a single reference is available, we use LivePortrait~\citep{guo2024liveportrait} to synthesize seven expressive references, driven by expressions from a held-out subject. The portrait and synthesized images form an eight-image gallery (see example in Figure~\ref{fig:liveportrait-gallery}, Appendix~\ref{sec:liveportrait-gallery}).

\section{Results}
\subsection{Qualitative Results}
\begin{figure}[htb]
    \centering
    \setlength{\tabcolsep}{1pt}
    \begin{tabular}{@{}c@{\hspace{0.2em}}c@{\hspace{0.2em}}cccc@{}}
        \smash{\raisebox{\dimexpr-0.5\height-0.11\linewidth\relax}{\shortstack{\includegraphics[height=0.17\linewidth]{fig/video_samples/reference_ZehaoLi_Cam10.png}\\[-0.25em]\scriptsize Neutral Face\\[0.2em]\includegraphics[height=0.17\linewidth]{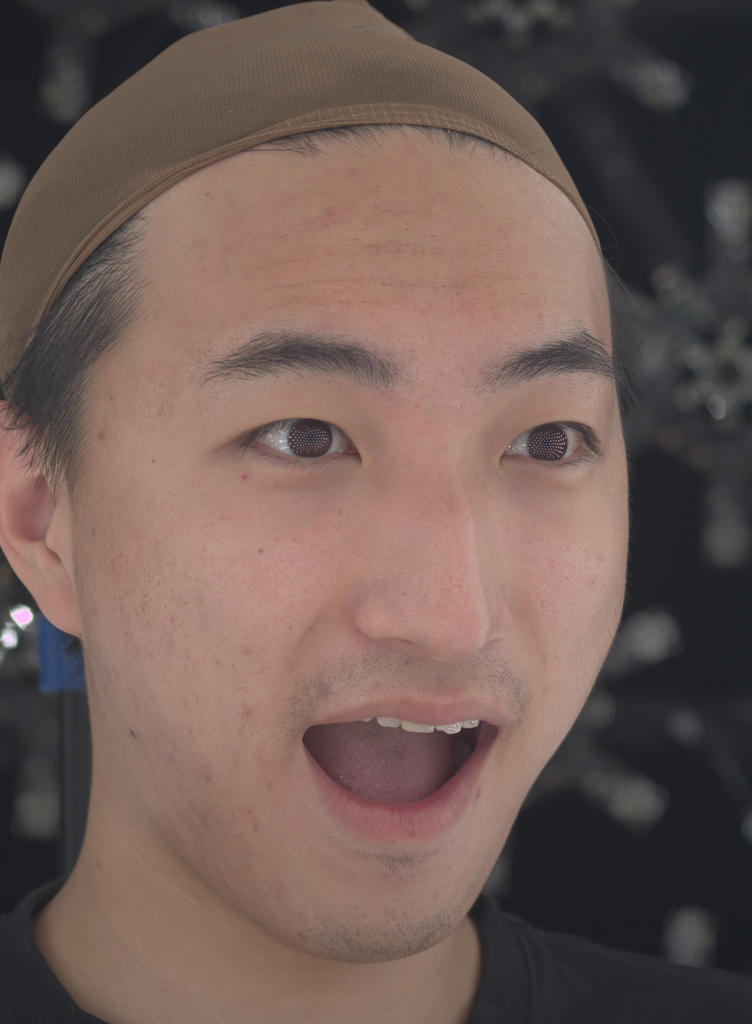}\\[-0.25em]\scriptsize Mouth Open}}} &
        \rotatebox{90}{\scriptsize AnyID} &
        \includegraphics[width=0.19\linewidth]{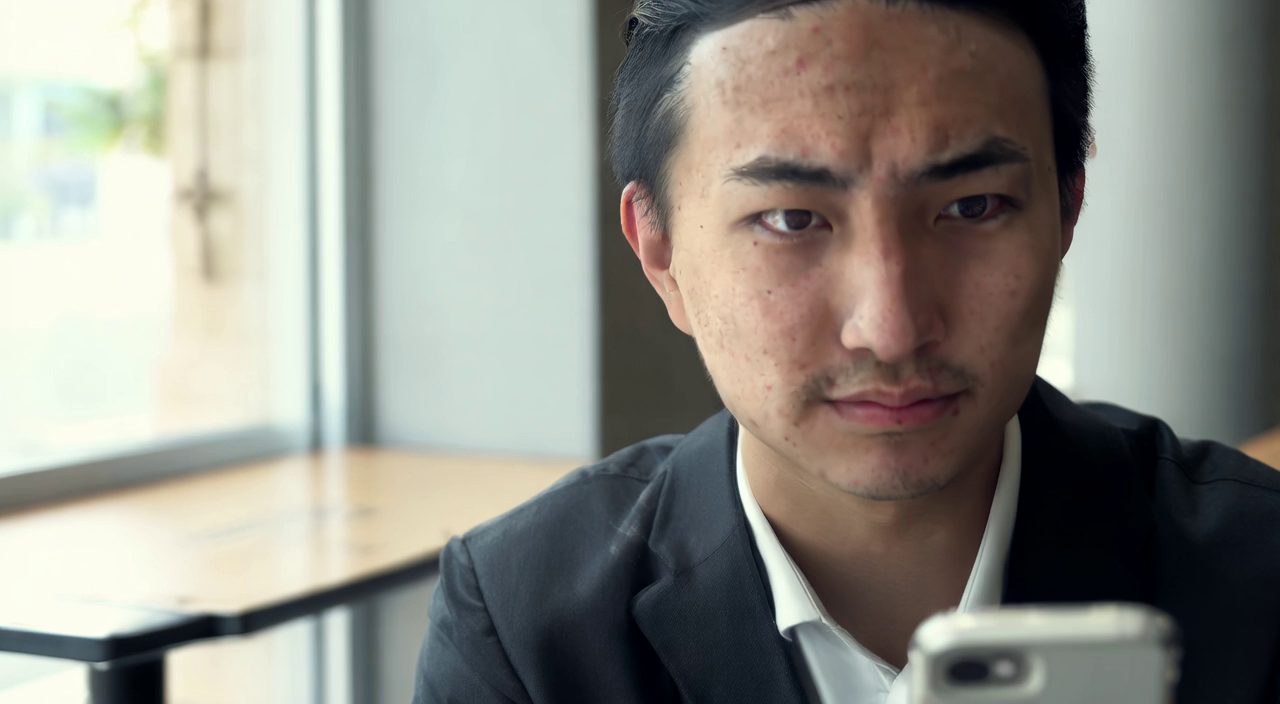} &
        \includegraphics[width=0.19\linewidth]{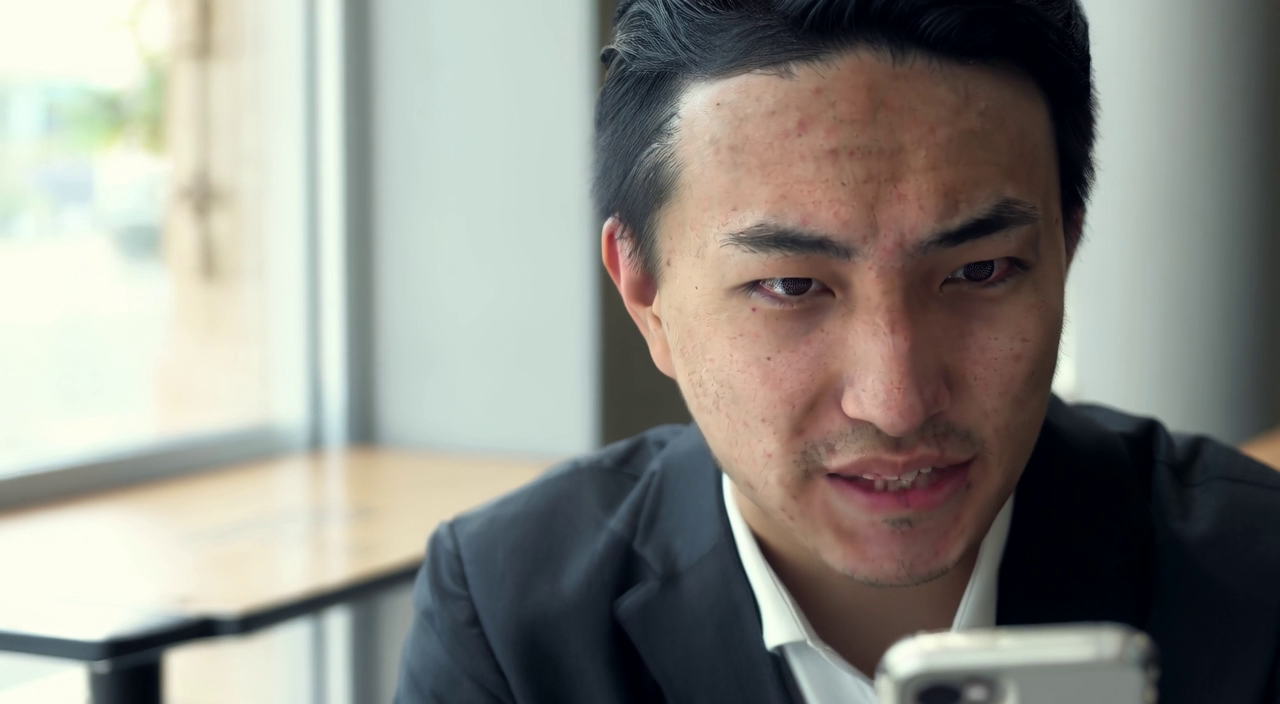} &
        \includegraphics[width=0.19\linewidth]{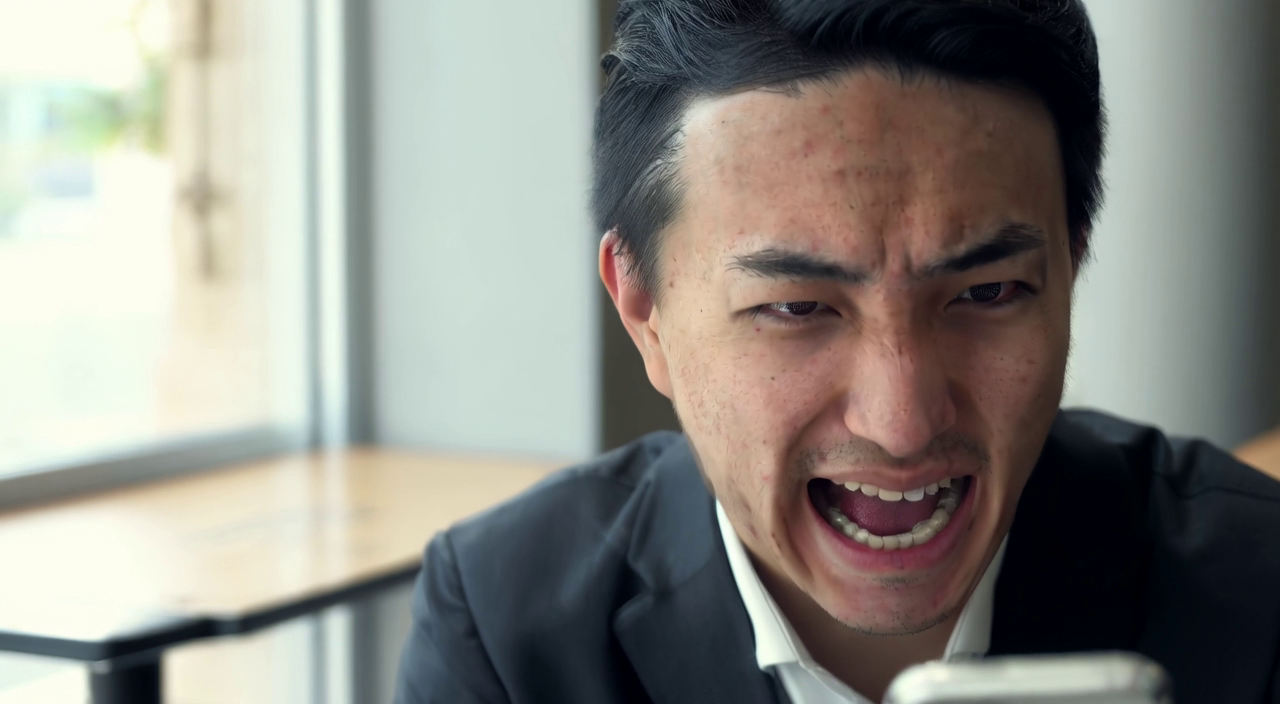} &
        \includegraphics[width=0.19\linewidth]{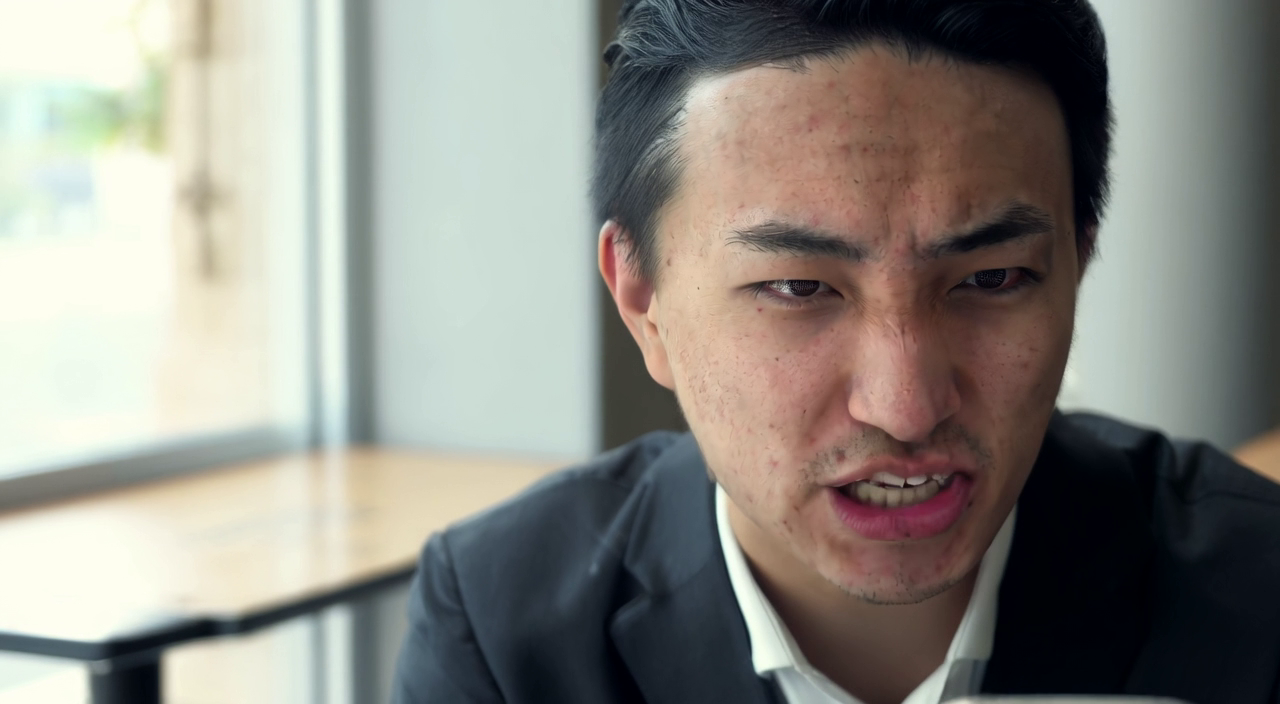} \\
        & \rotatebox{90}{\scriptsize Stand-In} &
        \includegraphics[width=0.19\linewidth]{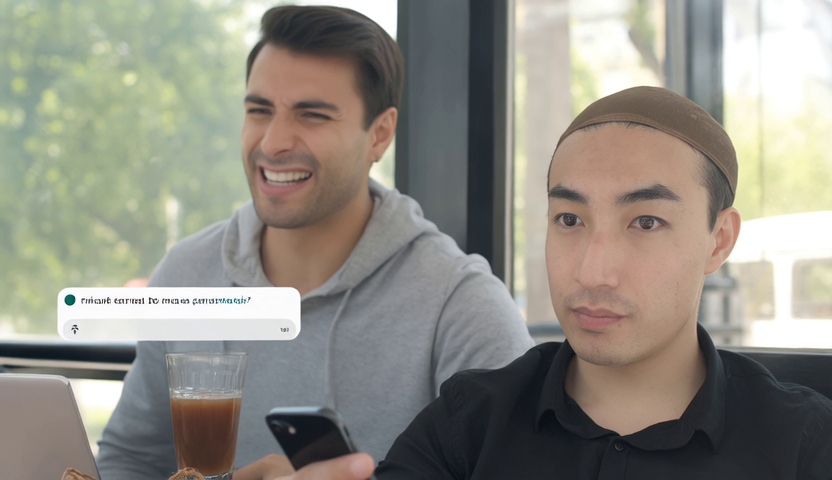} &
        \includegraphics[width=0.19\linewidth]{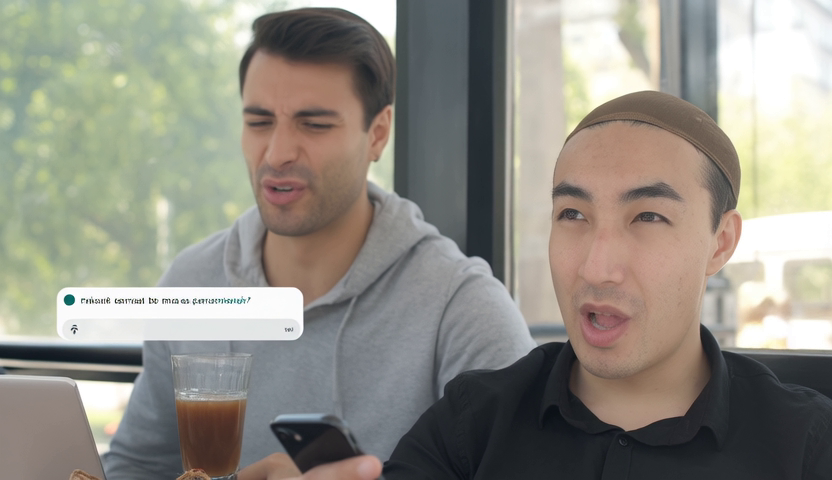} &
        \includegraphics[width=0.19\linewidth]{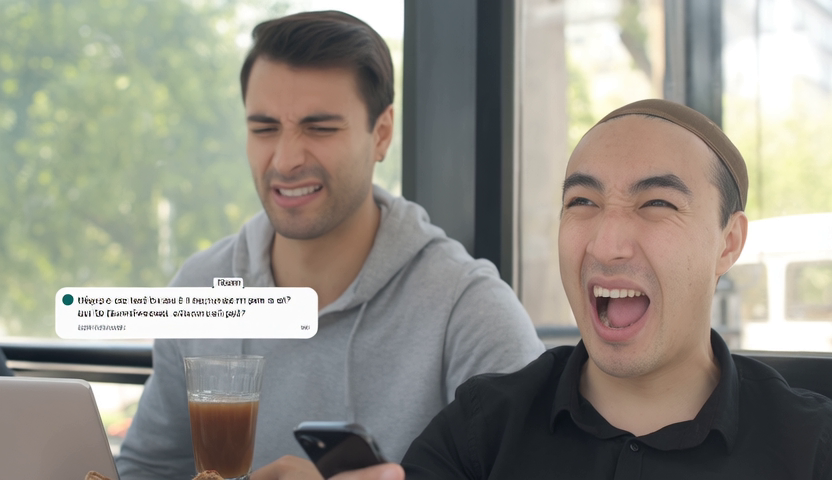} &
        \includegraphics[width=0.19\linewidth]{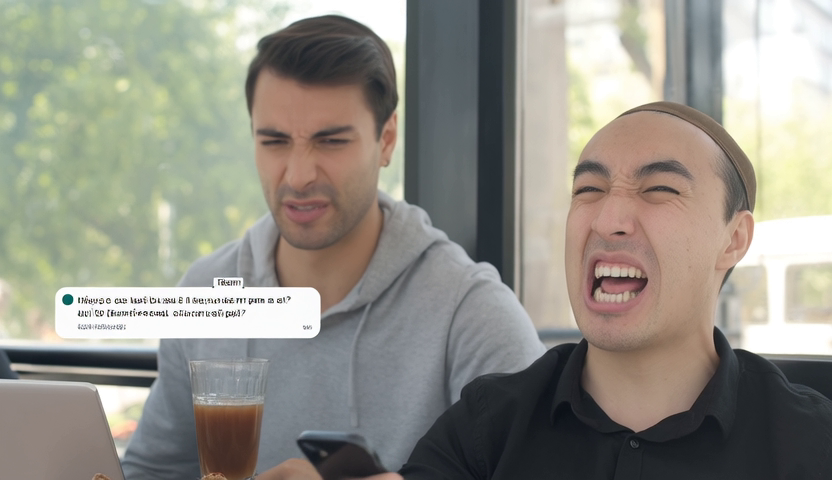} \\
        & \rotatebox{90}{\scriptsize Ours (Synthetic)} &
        \includegraphics[width=0.19\linewidth]{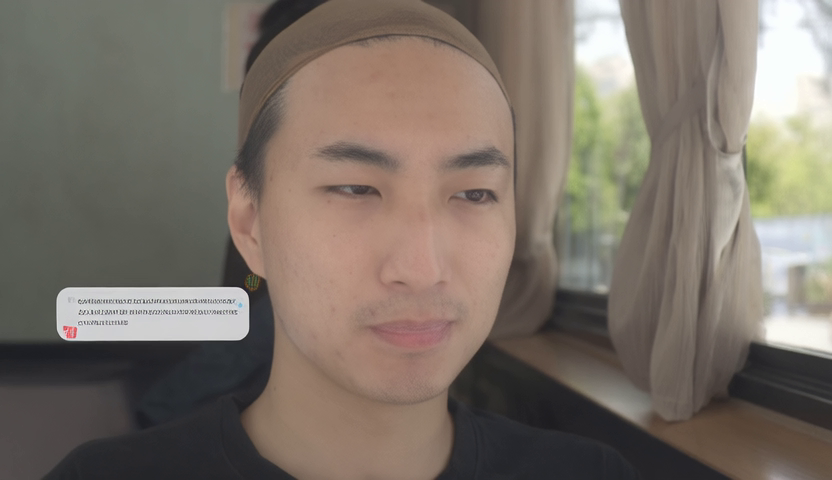} &
        \includegraphics[width=0.19\linewidth]{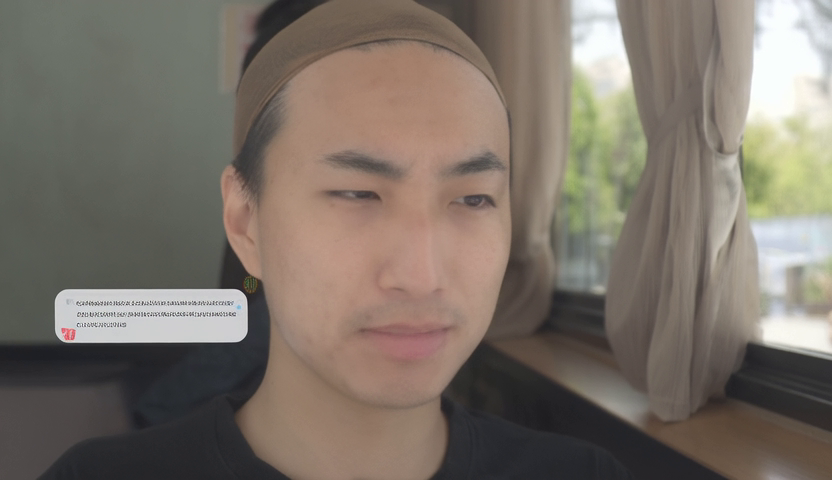} &
        \includegraphics[width=0.19\linewidth]{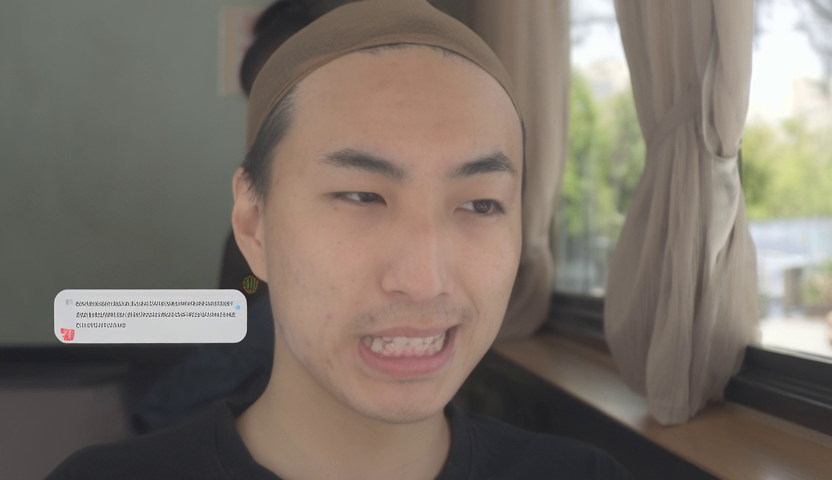} &
        \includegraphics[width=0.19\linewidth]{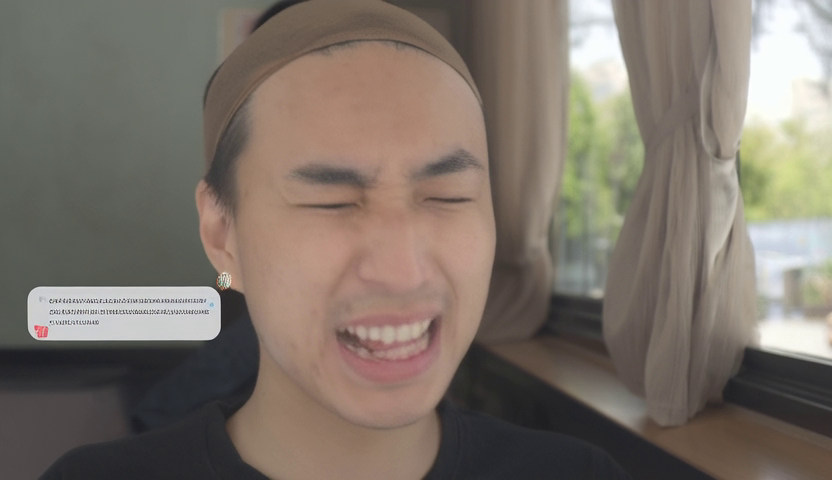} \\
        & \rotatebox{90}{\scriptsize Ours (Real)} &
        \includegraphics[width=0.19\linewidth]{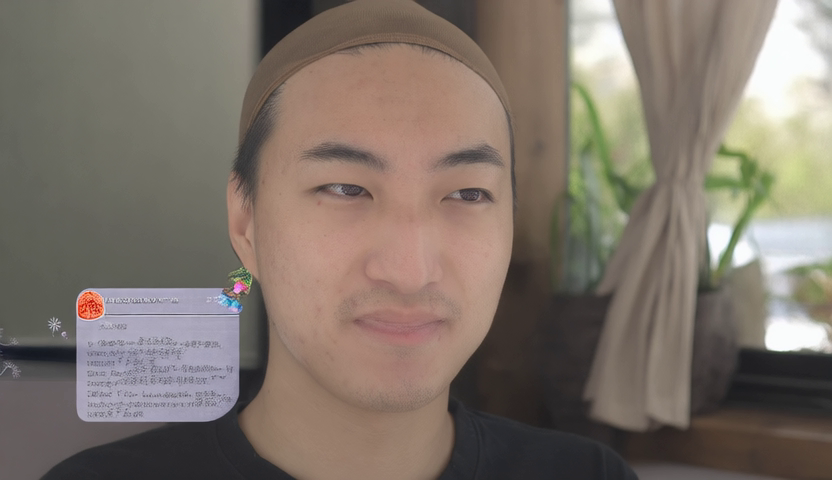} &
        \includegraphics[width=0.19\linewidth]{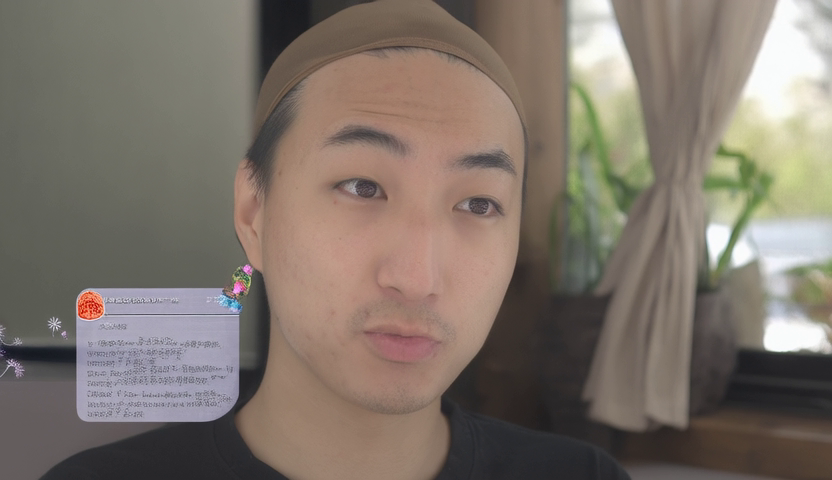} &
        \includegraphics[width=0.19\linewidth]{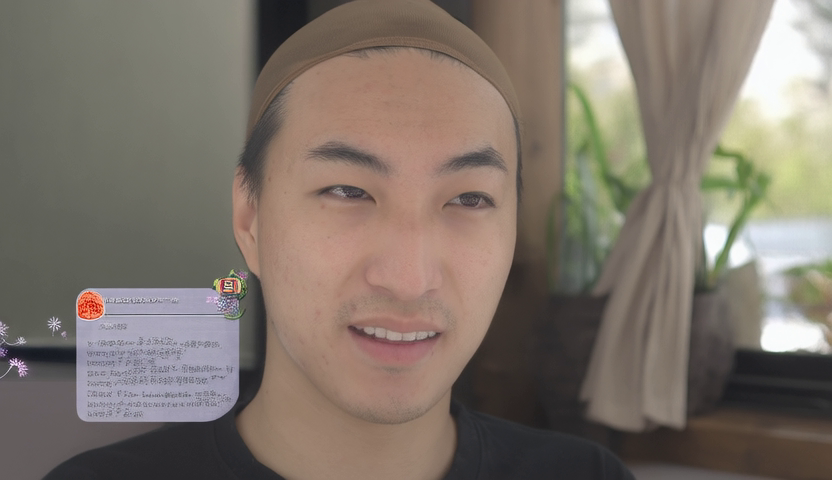} &
        \includegraphics[width=0.19\linewidth]{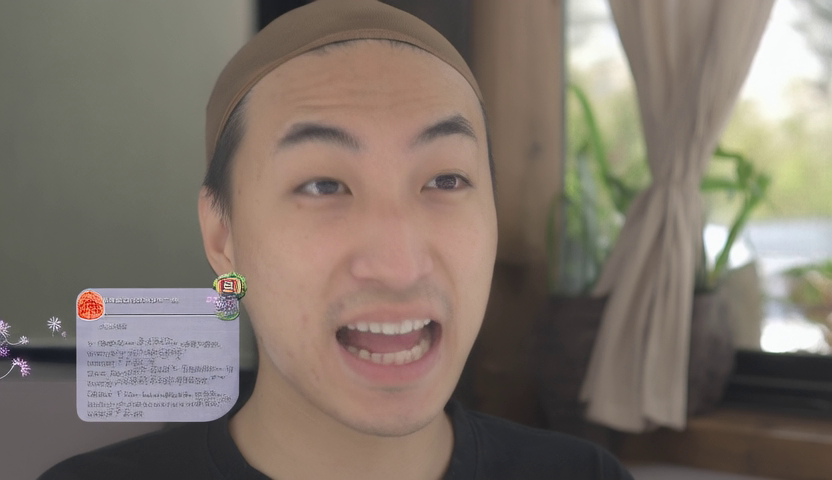} \\[0.4em]
        \smash{\raisebox{\dimexpr-0.5\height-0.11\linewidth\relax}{\shortstack{\includegraphics[height=0.17\linewidth]{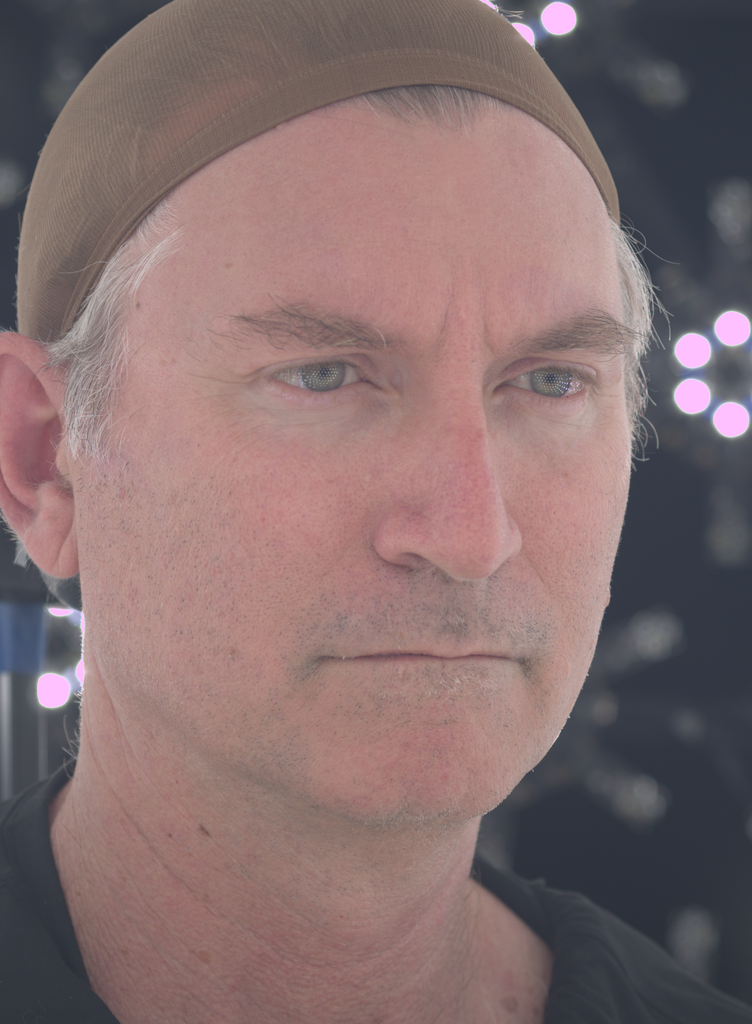}\\[-0.25em]\scriptsize Neutral Face\\[0.2em]\includegraphics[height=0.17\linewidth]{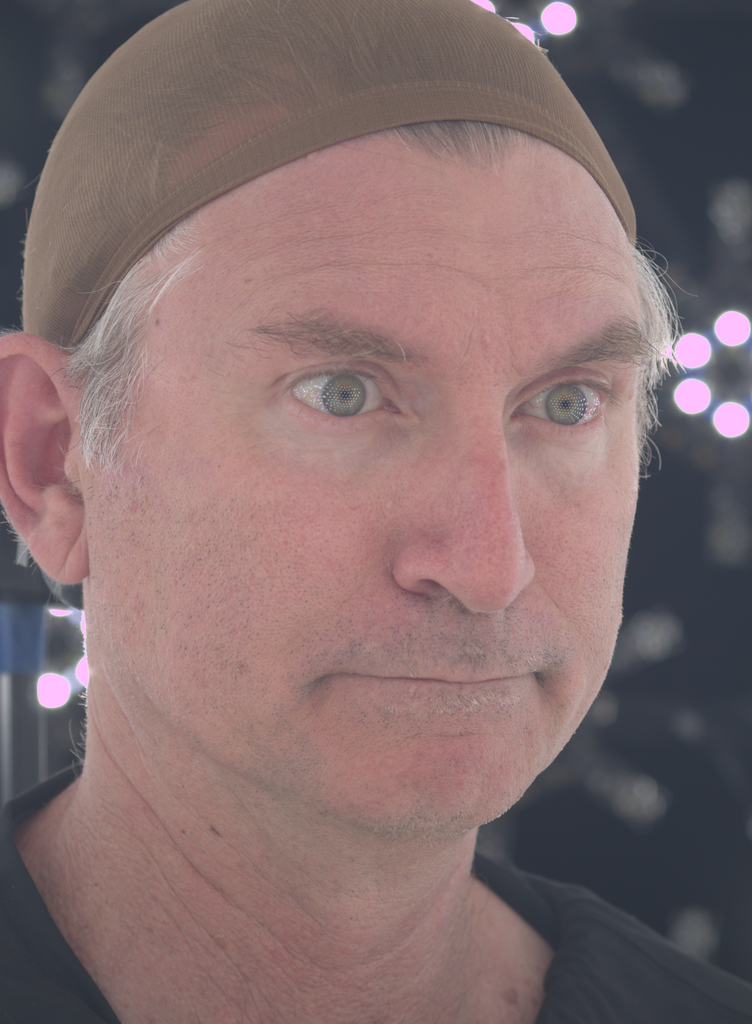}\\[-0.25em]\scriptsize Frown}}} &
        \rotatebox{90}{\scriptsize AnyID} &
        \includegraphics[width=0.19\linewidth]{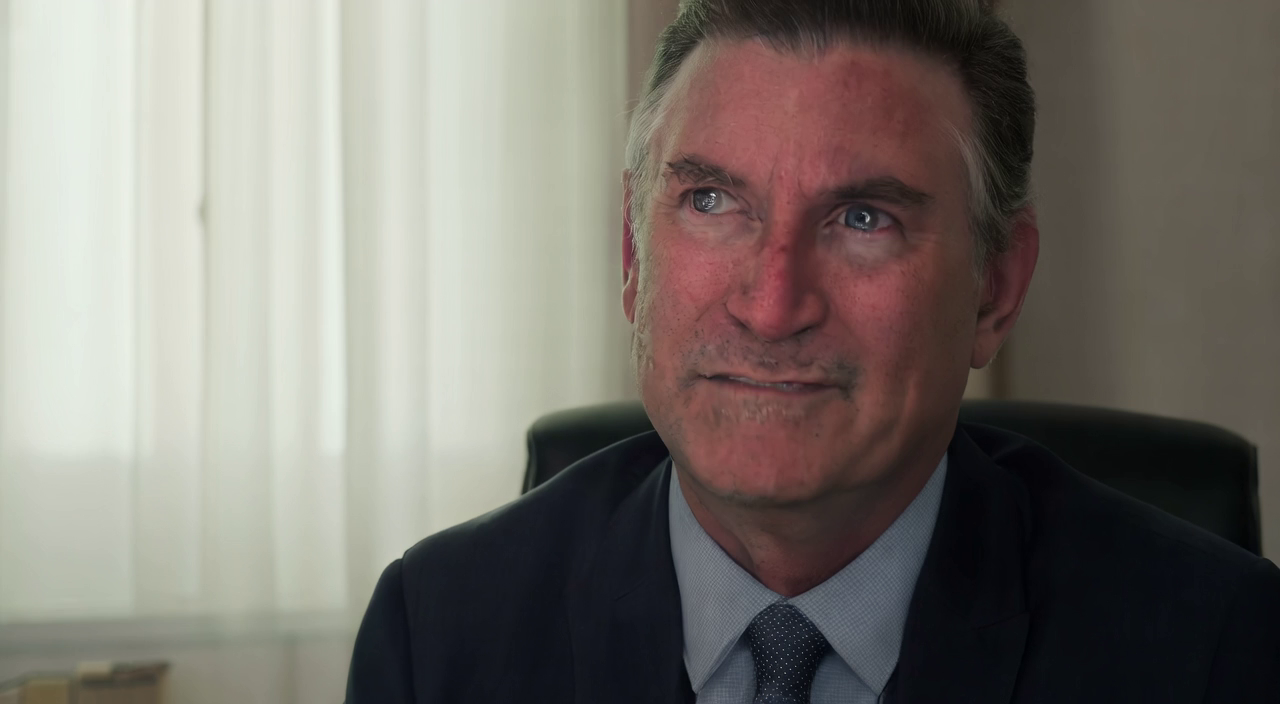} &
        \includegraphics[width=0.19\linewidth]{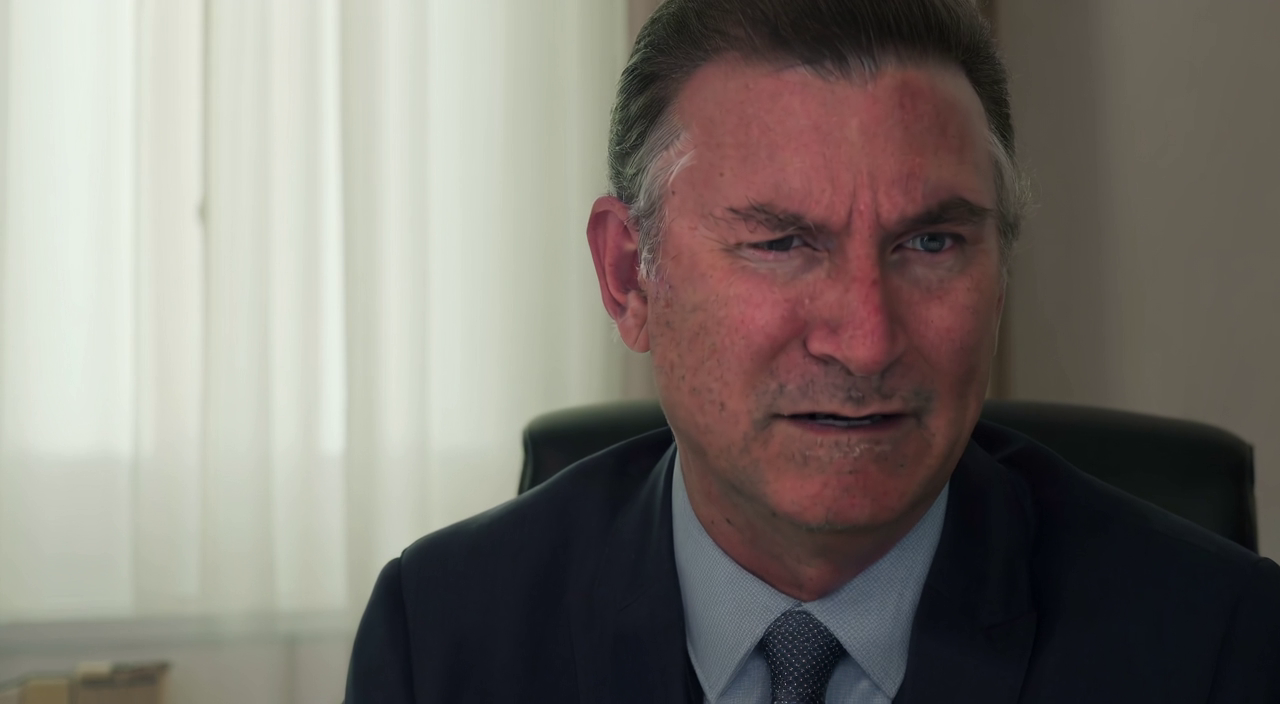} &
        \includegraphics[width=0.19\linewidth]{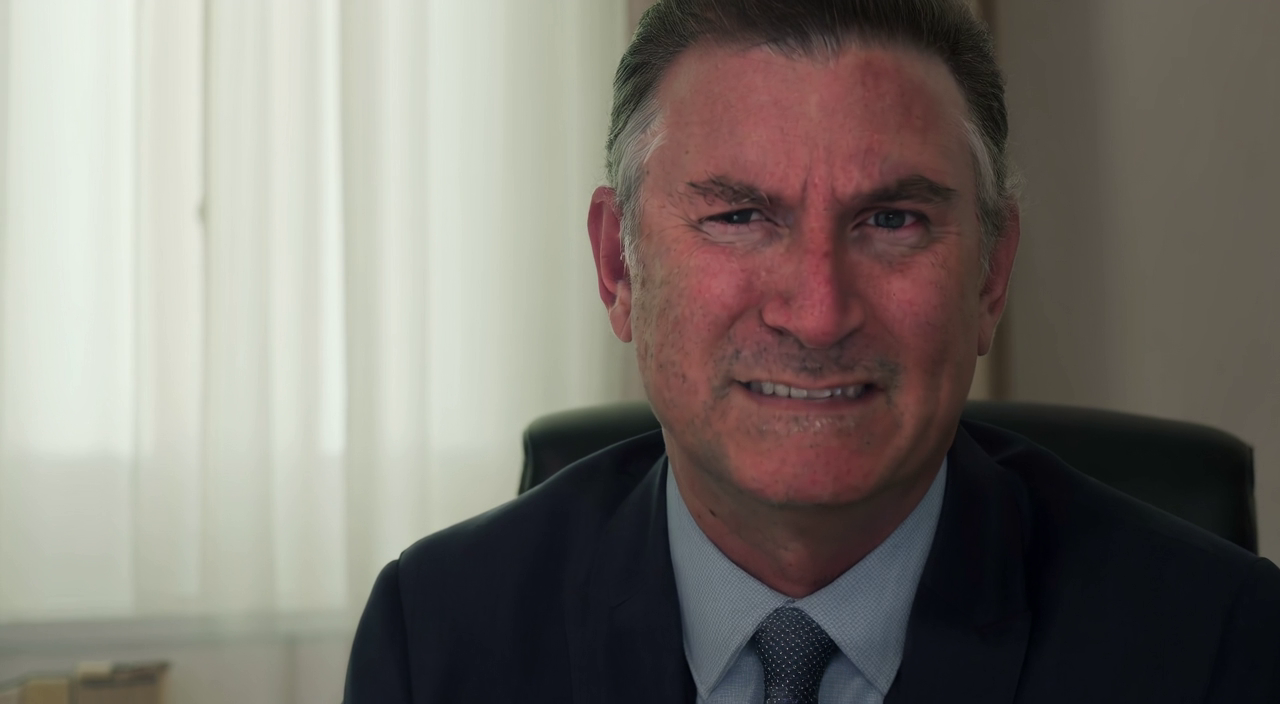} &
        \includegraphics[width=0.19\linewidth]{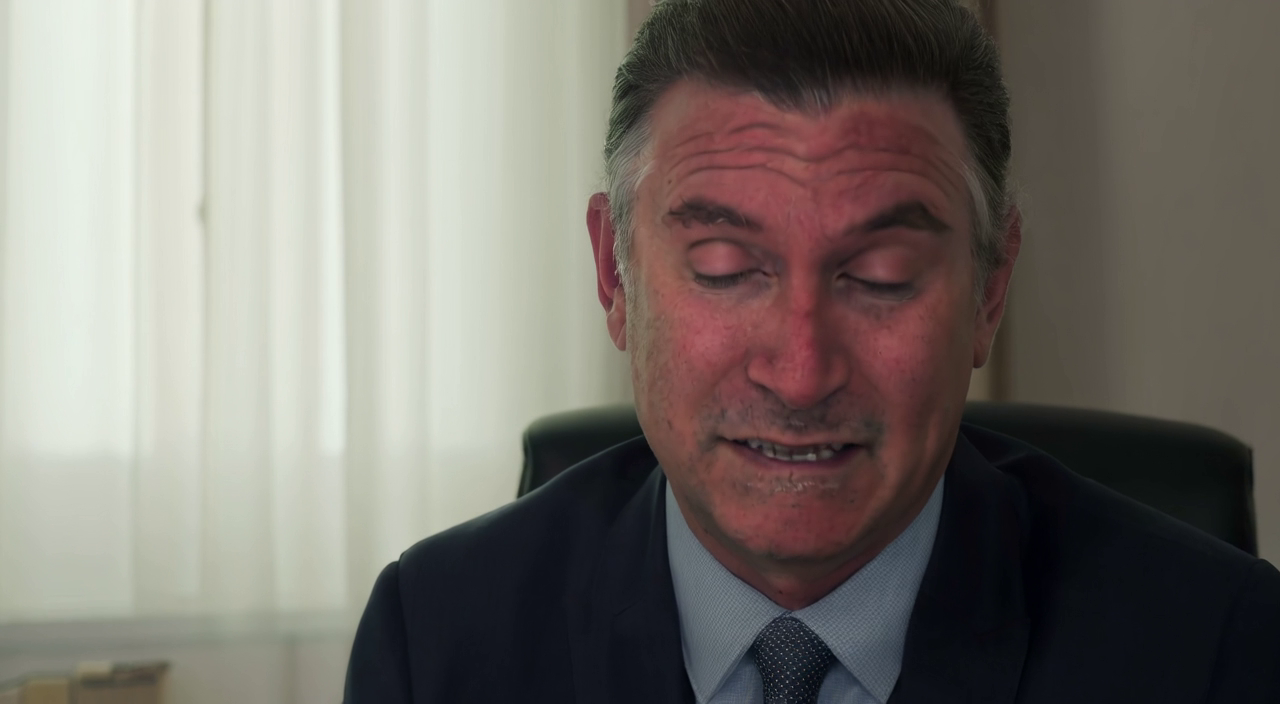} \\
        & \rotatebox{90}{\scriptsize Stand-In} &
        \includegraphics[width=0.19\linewidth]{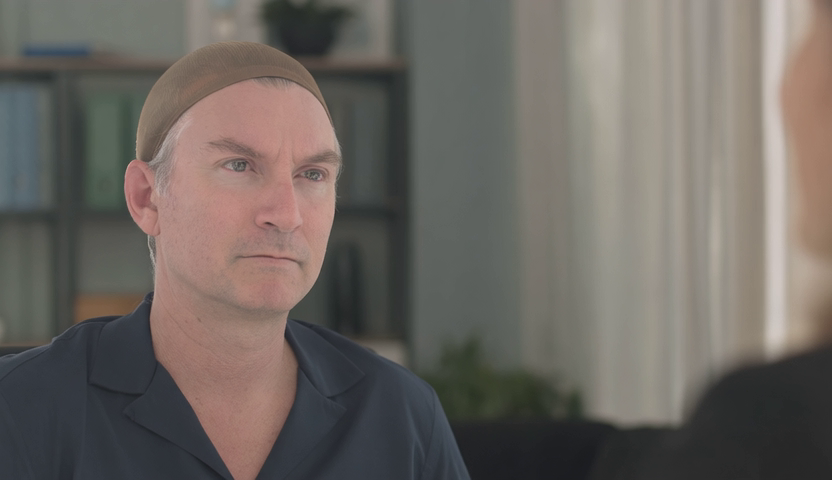} &
        \includegraphics[width=0.19\linewidth]{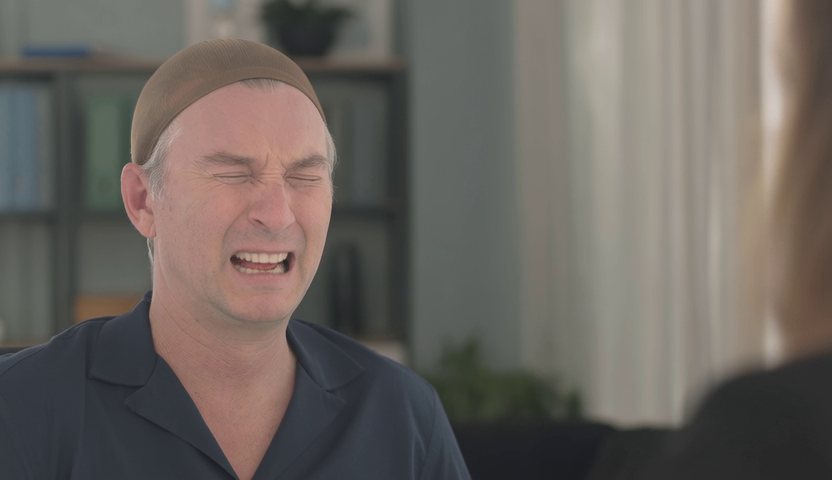} &
        \includegraphics[width=0.19\linewidth]{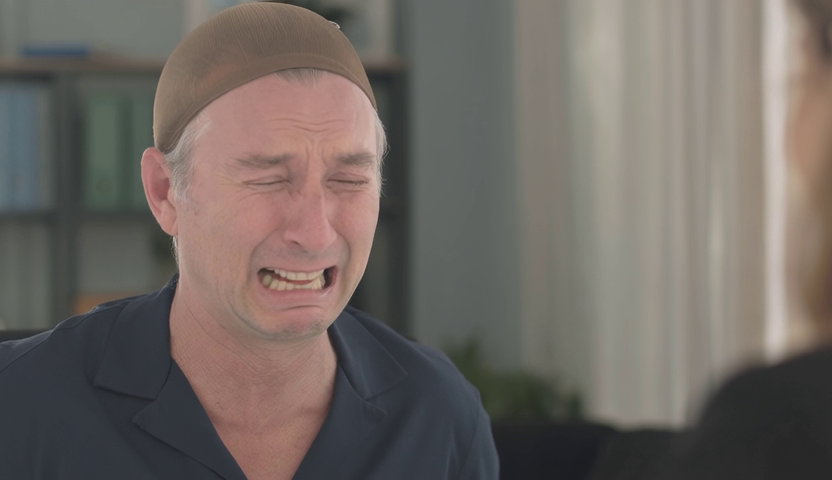} &
        \includegraphics[width=0.19\linewidth]{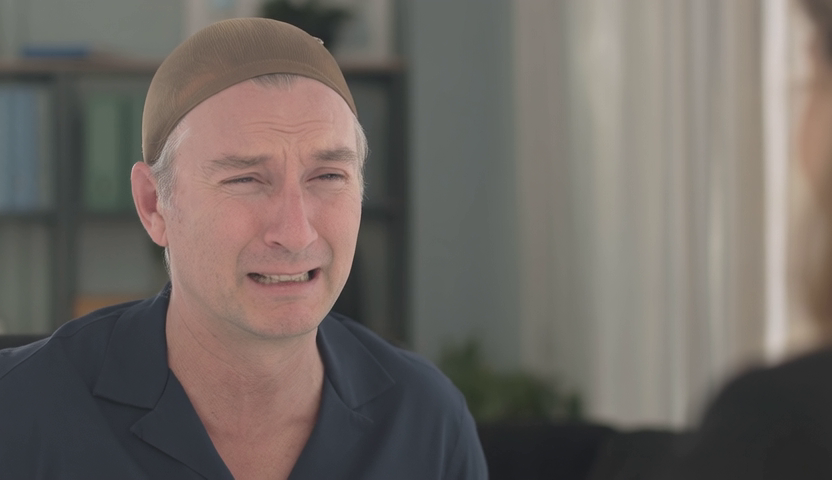} \\
        & \rotatebox{90}{\scriptsize Ours (Synthetic)} &
        \includegraphics[width=0.19\linewidth]{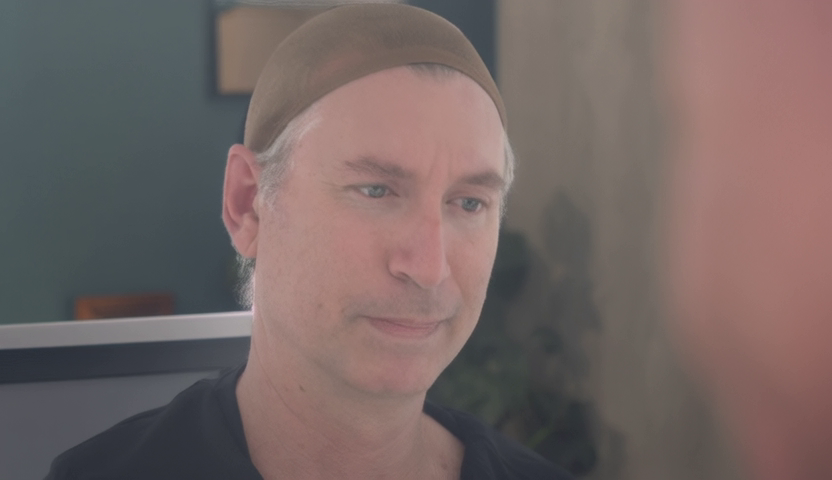} &
        \includegraphics[width=0.19\linewidth]{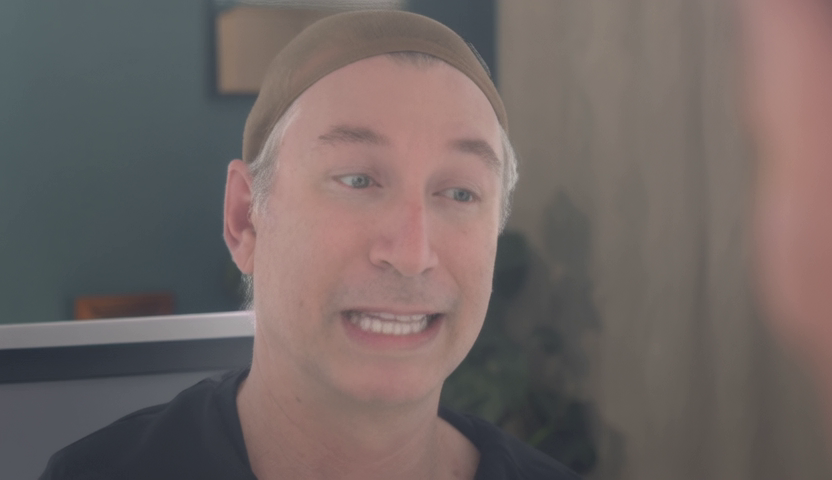} &
        \includegraphics[width=0.19\linewidth]{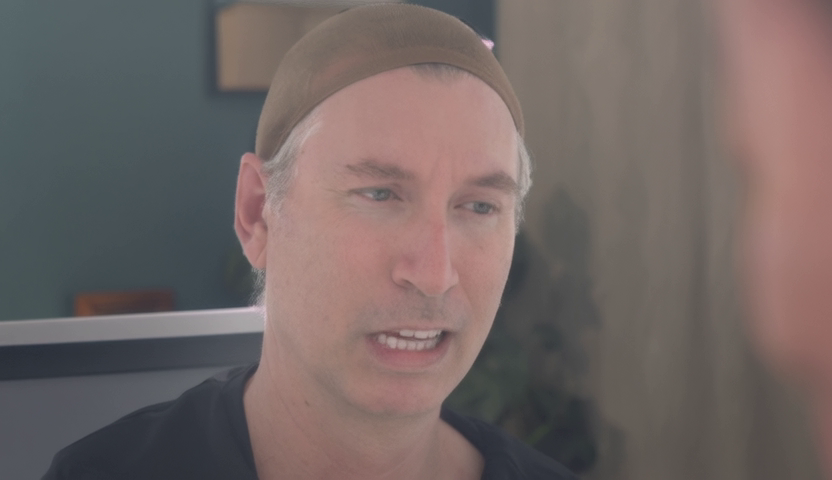} &
        \includegraphics[width=0.19\linewidth]{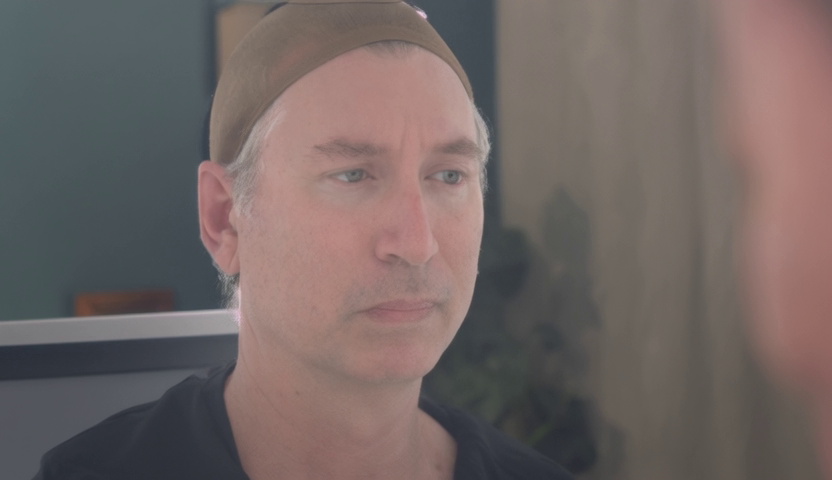} \\
        & \rotatebox{90}{\scriptsize Ours (Real)} &
        \includegraphics[width=0.19\linewidth]{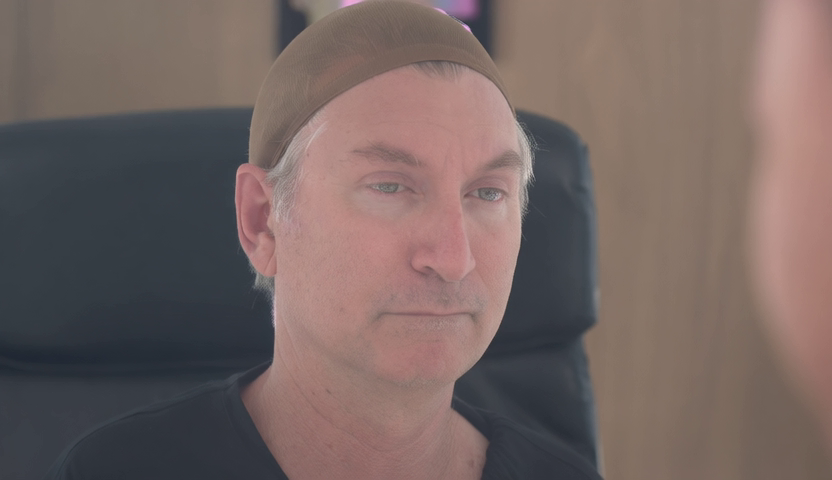} &
        \includegraphics[width=0.19\linewidth]{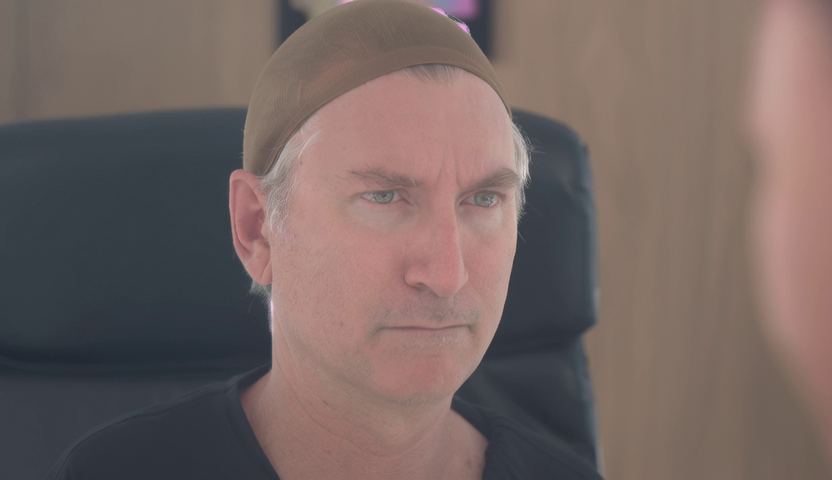} &
        \includegraphics[width=0.19\linewidth]{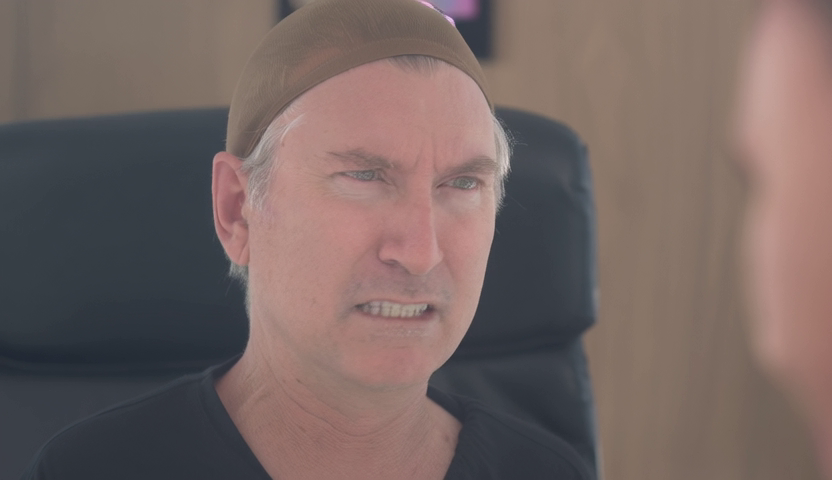} &
        \includegraphics[width=0.19\linewidth]{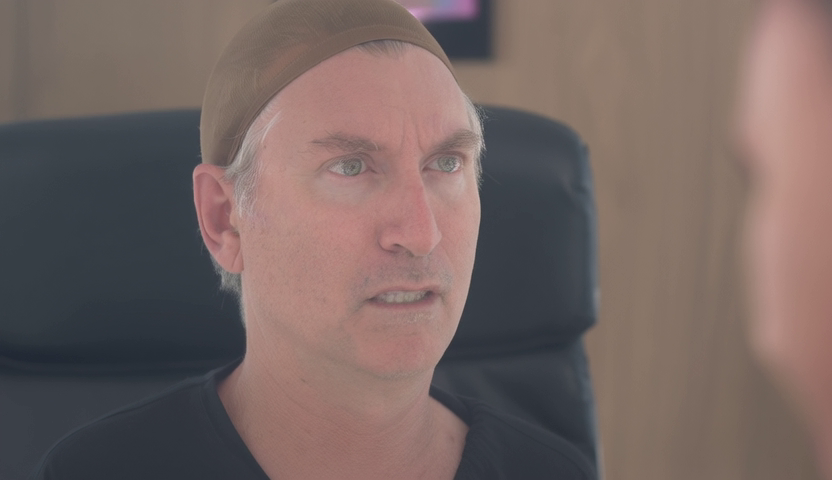}
    \end{tabular}
    \caption{Qualitative comparison. The leftmost images are the subject's neutral portrait and one expressive reference; each row shows four temporally ordered frames from the same generated video.}
    \label{fig:qualitative-results}
\end{figure}

In Figure~\ref{fig:qualitative-results}, we show videos generated by our model using a real reference gallery, \textit{Ours (Real)}, or a LivePortrait-generated gallery, \textit{Ours (Synthetic)}, alongside the representative baselines Stand-In and AnyID. With information from a diverse set of expressions, our model produces frames that better preserve identity under strong expressions.

\subsection{Results on Our Evaluation Dataset}
We evaluate the baselines and our models on the evaluation dataset introduced above. Table~\ref{tab:curated-identity-results} shows the nearest-gallery identity similarities $s_{\mathrm{Arc}}^*$ and $s_{\mathrm{Cur}}^*$, stratified by expression intensity. Our model, whether using synthetic or real reference galleries, outperforms all baselines in all three expression-intensity regimes. The performance gap is especially pronounced in the intense and extreme regimes, demonstrating the effectiveness of expression-diverse multi-reference conditioning in preserving identity under expressive motion.

\begin{table}[htbp]
    \centering
    \caption{Identity fidelity on our evaluation dataset. The percentage columns report the fraction of each model's frames in each expression-intensity regime.}
    \label{tab:curated-identity-results}
    \resizebox{\textwidth}{!}{%
    \begin{tabular}{lrrrrrrrrrrr}
        \toprule
        & \multicolumn{2}{c}{All frames} & \multicolumn{3}{c}{Mild} & \multicolumn{3}{c}{Intense} & \multicolumn{3}{c}{Extreme} \\
        \cmidrule(lr){2-3} \cmidrule(lr){4-6} \cmidrule(lr){7-9} \cmidrule(l){10-12}
        Model & $s_{\mathrm{Arc}}^* \uparrow$ & $s_{\mathrm{Cur}}^* \uparrow$ & \% & $s_{\mathrm{Arc}}^* \uparrow$ & $s_{\mathrm{Cur}}^* \uparrow$ & \% & $s_{\mathrm{Arc}}^* \uparrow$ & $s_{\mathrm{Cur}}^* \uparrow$ & \% & $s_{\mathrm{Arc}}^* \uparrow$ & $s_{\mathrm{Cur}}^* \uparrow$ \\
        \midrule
        Concat-ID [ICCVW2025] & 59.31 & 58.19 & 24.6 & 71.64 & 70.44 & 54.7 & 58.47 & 57.17 & 20.7 & 46.82 & 46.30 \\
        Phantom [ICCV2025] & 65.67 & 66.85 & 7.0 & 80.40 & 79.77 & 40.8 & 71.84 & 72.20 & 52.2 & 58.88 & 60.95 \\
        ConsisID [CVPR2025] & 33.02 & 31.81 & 3.0 & 50.88 & 50.53 & 50.3 & 39.19 & 38.31 & 46.8 & 25.24 & 23.63 \\
        Stand-In [CVPR2026] & 54.37 & 53.97 & 15.9 & 81.04 & 81.24 & 39.7 & 56.87 & 56.28 & 44.5 & 42.63 & 42.18 \\
        AnyID [CVPR2026] & 62.83 & 62.74 & 7.8 & 73.86 & 73.77 & 54.4 & 66.36 & 66.02 & 37.8 & 55.48 & 55.77 \\
        Gen-4.5 & 55.02 & 54.52 & 22.1 & 76.41 & 76.99 & 44.2 & 55.66 & 55.07 & 33.7 & 40.17 & 39.07 \\
        \midrule
        Ours (Synthetic) & 76.35 & 76.54 & 14.9 & 83.16 & 84.31 & 71.0 & 76.48 & 76.53 & 14.1 & 68.48 & 68.36 \\
        Ours (Real) & \textbf{80.78} & \textbf{81.68} & 29.0 & \textbf{83.68} & \textbf{85.03} & 63.5 & \textbf{80.23} & \textbf{80.95} & 7.4 & \textbf{74.27} & \textbf{74.80} \\
        \bottomrule
    \end{tabular}%
    }
\end{table}

Furthermore, Table~\ref{tab:drive-matched-identity-wins}
shows that our model achieves higher identity fidelity on matched subject--prompt pairs, including the subset where our real-gallery model produces greater peak expression intensity than the competing method.

\begin{table}[htb]
    \centering
    \caption{Win rates for Ours (Real) on matched subject--prompt pairs, using per-video mean nearest-gallery similarity. The right-hand columns include only pairs where ours has greater peak expression intensity.}
    \label{tab:drive-matched-identity-wins}
    \resizebox{0.65\textwidth}{!}{%
    \begin{tabular}{lrrrr}
        \toprule
        & \multicolumn{2}{c}{All pairs} & \multicolumn{2}{c}{Ours has higher peak intensity} \\
        \cmidrule(lr){2-3} \cmidrule(l){4-5}
        Baseline & $s_{\mathrm{Arc}}^* \uparrow$ & $s_{\mathrm{Cur}}^* \uparrow$ & $s_{\mathrm{Arc}}^* \uparrow$ & $s_{\mathrm{Cur}}^* \uparrow$ \\
        \midrule
        Concat-ID & 97.9\% & 99.0\% & 96.1\% & 98.3\% \\
        Phantom & 98.8\% & 98.5\% & 96.9\% & 96.9\% \\
        ConsisID & 100.0\% & 100.0\% & 100.0\% & 100.0\% \\
        Stand-In & 97.9\% & 98.3\% & 96.4\% & 96.4\% \\
        AnyID & 99.0\% & 99.2\% & 96.9\% & 97.7\% \\
        Gen-4.5 & 100.0\% & 100.0\% & 100.0\% & 100.0\% \\
        \bottomrule
    \end{tabular}
    }
\end{table}

\section{Ablation Studies}
\subsection{Number of Reference Images}
Increasing the reference count improves overall CurricularFace similarity by approximately $1.3$ points per added reference, with the largest gains under intense and extreme expressions (Table~\ref{tab:reference-count-ablation}). A ninth reference provides no further overall gain, which justifies our choice of using 8 references in the gallery. For context, we also report AnyID~\citep{Wang_2026_CVPR}, which accepts 5 references.

\begin{table}[htb]
    \centering
    \caption{Effect of reference count on identity fidelity. Each additional reference adds an expression to the preceding reference set; the ninth reference uses an held-out expression outside the standard gallery. Scores within 0.5 points of the best value in each similarity column are bolded.}
    \label{tab:reference-count-ablation}
    \resizebox{\textwidth}{!}{%
    \begin{tabular}{lrrrrrrrrrrrr}
        \toprule
        & & \multicolumn{2}{c}{All frames} & \multicolumn{3}{c}{Mild} & \multicolumn{3}{c}{Intense} & \multicolumn{3}{c}{Extreme} \\
        \cmidrule(lr){3-4} \cmidrule(lr){5-7} \cmidrule(lr){8-10} \cmidrule(l){11-13}
        Model & Refs. & $s_{\mathrm{Arc}}^* \uparrow$ & $s_{\mathrm{Cur}}^* \uparrow$ & \% & $s_{\mathrm{Arc}}^* \uparrow$ & $s_{\mathrm{Cur}}^* \uparrow$ & \% & $s_{\mathrm{Arc}}^* \uparrow$ & $s_{\mathrm{Cur}}^* \uparrow$ & \% & $s_{\mathrm{Arc}}^* \uparrow$ & $s_{\mathrm{Cur}}^* \uparrow$ \\
        \midrule
        AnyID & 5 & 62.83 & 62.74 & 7.8 & 73.86 & 73.77 & 54.4 & 66.36 & 66.02 & 37.8 & 55.48 & 55.77 \\
        \midrule
        Ours & 5 & 77.22 & 77.77 & 24.9 & \textbf{83.92} & \textbf{85.05} & 60.6 & 77.03 & 77.39 & 14.5 & 66.51 & 66.82 \\
        Ours & 6 & 78.36 & 78.94 & 25.3 & \textbf{83.62} & \textbf{84.78} & 63.8 & 77.83 & 78.24 & 10.8 & 69.17 & 69.39 \\
        Ours & 7 & 79.59 & 80.34 & 27.4 & \textbf{83.72} & \textbf{85.03} & 63.5 & 78.93 & 79.49 & 9.1 & 71.80 & 72.18 \\
        Ours & 8 & \textbf{80.78} & \textbf{81.68} & 29.0 & \textbf{83.68} & \textbf{85.03} & 63.5 & \textbf{80.23} & \textbf{80.95} & 7.4 & \textbf{74.27} & \textbf{74.80} \\
        Ours & 9 & \textbf{80.71} & \textbf{81.45} & 28.4 & \textbf{83.45} & 84.51 & 64.6 & \textbf{80.33} & \textbf{80.97} & 7.1 & 73.16 & 73.50 \\
        \bottomrule
    \end{tabular}%
    }
\end{table}

\subsection{Expression Diversity vs. Pose Diversity}
Table~\ref{tab:reference-diversity-ablation} isolates the effect of expression diversity by comparing three sets of eight references: repeated neutral images, pose-diverse neutral images, and expression-diverse images. The results show that the expression-diverse set achieves higher fidelity than the pose-diverse set in all three regimes. The repeated neutral set achieves the highest fidelity in the mild regime, but shows a similar score in the intense regime and a lower score in the extreme regime compared to the expression-diverse set. The repeated neutral set also produces a much lower share of intense and extreme frames. This suggests that its high overall similarity comes partly from suppressing expression intensity, highlighting the importance of evaluation stratified by intensity.
\begin{table}[htbp]
    \centering
    \caption{Effect of reference-image diversity on identity similarity. Scores within 0.5 points of the best value in each similarity column are bolded.}
    \label{tab:reference-diversity-ablation}
    \resizebox{\textwidth}{!}{%
    \begin{tabular}{lrrrrrrrrrrr}
        \toprule
        & \multicolumn{2}{c}{All frames} & \multicolumn{3}{c}{Mild} & \multicolumn{3}{c}{Intense} & \multicolumn{3}{c}{Extreme} \\
        \cmidrule(lr){2-3} \cmidrule(lr){4-6} \cmidrule(lr){7-9} \cmidrule(l){10-12}
        Reference set & $s_{\mathrm{Arc}}^* \uparrow$ & $s_{\mathrm{Cur}}^* \uparrow$ & \% & $s_{\mathrm{Arc}}^* \uparrow$ & $s_{\mathrm{Cur}}^* \uparrow$ & \% & $s_{\mathrm{Arc}}^* \uparrow$ & $s_{\mathrm{Cur}}^* \uparrow$ & \% & $s_{\mathrm{Arc}}^* \uparrow$ & $s_{\mathrm{Cur}}^* \uparrow$ \\
        \midrule
        Repeated neutral & \textbf{85.85} & \textbf{86.36} & 71.5 & \textbf{88.21} & \textbf{88.71} & 27.3 & \textbf{80.55} & \textbf{81.06} & 1.1 & 64.93 & 66.31 \\
        Pose-diverse neutral & 57.27 & 57.41 & 8.7 & 67.75 & 67.71 & 60.0 & 59.46 & 59.52 & 31.2 & 50.11 & 50.47 \\
        Expression-diverse & 80.78 & 81.68 & 29.0 & 83.68 & 85.03 & 63.5 & \textbf{80.23} & \textbf{80.95} & 7.4 & \textbf{74.27} & \textbf{74.80} \\
        \bottomrule
    \end{tabular}%
    }
\end{table}

\subsection{Temporal Positional Embedding}
We design our network to treat the reference images as an unordered set, so all the references are assigned the same temporal embedding to avoid introducing an ordering bias during training. For comparison, we also train a variant that assigns each reference a distinct temporal embedding. During training, we shuffle the reference images, and for each image $i=1,2,\cdots,N$ in the shuffled order, we assign it a temporal embedding $-(i+1)$ on the video-to-image attention path. Reference-image self-attention retains the shared coordinate $-1$. This negative-temporal variant is trained and evaluated with the same eight expression-diverse references and generation settings as ours. The shared-coordinate model achieves higher identity fidelity in all regimes, shown in Table~\ref{tab:temporal-position-ablation}.

\begin{table}[htb]
    \centering
    \caption{Effect of temporal positional embeddings for multiple references.}
    \label{tab:temporal-position-ablation}
    \resizebox{\textwidth}{!}{%
    \begin{tabular}{lrrrrrrrrrrr}
        \toprule
        & \multicolumn{2}{c}{All frames} & \multicolumn{3}{c}{Mild} & \multicolumn{3}{c}{Intense} & \multicolumn{3}{c}{Extreme} \\
        \cmidrule(lr){2-3} \cmidrule(lr){4-6} \cmidrule(lr){7-9} \cmidrule(l){10-12}
        Temporal coordinates & $s_{\mathrm{Arc}}^* \uparrow$ & $s_{\mathrm{Cur}}^* \uparrow$ & \% & $s_{\mathrm{Arc}}^* \uparrow$ & $s_{\mathrm{Cur}}^* \uparrow$ & \% & $s_{\mathrm{Arc}}^* \uparrow$ & $s_{\mathrm{Cur}}^* \uparrow$ & \% & $s_{\mathrm{Arc}}^* \uparrow$ & $s_{\mathrm{Cur}}^* \uparrow$ \\
        \midrule
        Distinct negative & 79.98 & 80.84 & 31.9 & 82.75 & 84.20 & 61.3 & 79.35 & 79.96 & 6.7 & 72.56 & 72.94 \\
        Shared $t=-1$ (ours) & \textbf{80.78} & \textbf{81.68} & 29.0 & \textbf{83.68} & \textbf{85.03} & 63.5 & \textbf{80.23} & \textbf{80.95} & 7.4 & \textbf{74.27} & \textbf{74.80} \\
        \bottomrule
    \end{tabular}%
    }
\end{table}

\section{Conclusion}
We show the limitations of a single neutral portrait for capturing person-specific facial changes and evaluating identity under strong expressions. To address this, we introduce an expression-diverse gallery for evaluation and propose a multi-reference conditioning model, which improves identity similarity across all expression-intensity regimes, especially under extreme expressions. These results highlight the importance of expression diversity in identity-preserving video generation. We encourage future work to provide larger expression-diverse datasets for benchmarking and training.

\clearpage
\subsection*{AI use statement}

In this work, we used generative AI tools to provide feedback on research methodology and experiments and to help implement methods.
We have not used generative AI tools for proposing or refining hypotheses, cleaning and reformatting datasets, or interpreting results.
Generating synthetic datasets, helping develop theoretical models or conceptual frameworks, formulating mathematical claims, providing critical ingredients for proving mathematical claims, assisting in the writing of proofs, assisting with translation, and supporting qualitative and thematic data analysis are not applicable to this work.
Additionally, we used generative AI tools for proposing the title and editing the paper to improve readability. We have reviewed all AI-assisted work. AI-assisted edits are manually reviewed before incorporation into the submitted work. We evaluate feedback from generative AI and decide which suggestions to adopt. AI-written code is manually audited. We take responsibility for the final content of this work, including text, claims, or artifacts produced with the aid of generative AI.

\subsection*{Ethics statement}
Our work includes photographs of human faces and expressions. All subjects have signed consent forms, granting permission for the use of their images in our research and for publication in academic venues. We ensure that all data is handled in accordance with relevant privacy laws and regulations, and that any sensitive information is properly anonymized when necessary.

\subsection*{Reproducibility statement}
We support reproducibility by providing detailed descriptions of our evaluation pipelines and training and inference settings, along with the expressions and prompts used in our evaluation set, in the paper and supplementary materials. We will make code, data, and model weights publicly available upon acceptance, before the camera-ready version is submitted.

\bibliography{citations}
\bibliographystyle{iclr2027_conference}

\clearpage
\appendix
\section{Implementation Details}
\label{sec:implementation-details}


\subsection{Face Detection and Expression Intensity Definition}
\label{sec:face-pipeline}

We detect the faces with SCRFD-10g~\citep{guo2021sample} at a detection size of $640\times640$ with a confidence threshold of $0.5$. When more than one face is found we keep the one with the highest detection confidence. The detector's five facial keypoints define a similarity transform to the standard ArcFace template, giving a $112\times112$ aligned crop; this single crop is embedded by both recognizers, ArcFace~\citep{deng2019arcface} and CurricularFace~\citep{huang2020curricularface}. Identity similarity is the cosine similarity between $\ell_2$-normalized embeddings, reported $\times 100$, following the convention of \citet{Wang_2026_CVPR}. Frames with no detectable face are dropped from all frame-level statistics, and a video with no detectable face in any sampled frame is excluded from the paired comparisons. The share of sampled frames with a detected face is above $98.8\%$ for every evaluated method and above $99.7\%$ for ours. 

For expressions, the crops are encoded by the SMIRK~\cite{SMIRK:CVPR:2024} encoder. We define $\Psi = [\,\psi_{\mathrm{expr}};\; \psi_{\mathrm{eyelid}};\; \psi_{\mathrm{jaw}}\,]$, a $55$-dimensional vector formed by the $50$ FLAME~\citep{li2017learning} expression coefficients, the $2$ eyelid parameters, and the $3$ jaw-rotation parameters predicted by SMIRK. The expression intensity of a frame is $\Delta\Psi = \|\Psi - \Psi_{\mathrm{neutral}}\|_2$.

\subsection{MEAD Dataset Preprocessing}
\label{sec:mead-processing}
For MEAD (Figure~\ref{fig:expression-identity-mead}), we use frontal-view videos of 48 subjects covering seven non-neutral emotions, three intensity levels, and three utterances per emotion and intensity level. We sample eight evenly spaced frames per clip, yielding 24,192 frames; 24,190 have valid expression estimates and contribute to the correlation analysis. The neutral expression vector is the coordinate-wise median over sampled neutral-clip frames, and the neutral identity embedding is the normalized mean of their embeddings.

\section{Supplemental Experiments}
\subsection{Evaluation with Held-Out Expressions}
\label{sec:heldout-expression-evaluation}
We conduct two additional evaluations using four held-out expressions (H1, H2, H3, H4) that were not used in the original eight-image gallery. Each expression is available for all 10 evaluation subjects to form a 40-image held-out gallery.

Table~\ref{tab:heldout-expression-metric} shows the superiority of nearest-gallery matching over using only the neutral portrait for identity similarity evaluation across all held-out expressions. Nearest-gallery matching reduces the range of the four expression-level means from 3.81 to 0.70 for ArcFace and from 3.53 to 1.75 for CurricularFace. Thus, the benefit generalizes beyond the expressions represented in the gallery.

\begin{table}[htbp]
    \centering
    \caption{Identity similarity of real photographs showing expressions excluded from the eight-image gallery (mean $\pm$ standard error over 10 subjects).}
    \label{tab:heldout-expression-metric}
    \resizebox{0.88\textwidth}{!}{%
    \begin{tabular}{lrrrr}
        \toprule
        \multirow{2}{*}{Held-out expression} & \multicolumn{2}{c}{ArcFace} & \multicolumn{2}{c}{CurricularFace} \\
        \cmidrule(lr){2-3} \cmidrule(l){4-5}
        & Neutral & Nearest gallery & Neutral & Nearest gallery \\
        \midrule
        H1 & $78.05 \pm 3.35$ & $80.51 \pm 3.05$ & $78.70 \pm 3.76$ & $81.03 \pm 3.32$ \\
        H2 & $75.35 \pm 3.53$ & $80.94 \pm 3.59$ & $76.89 \pm 3.43$ & $82.33 \pm 3.34$ \\
        H3 & $77.85 \pm 2.79$ & $81.05 \pm 2.50$ & $79.96 \pm 2.17$ & $82.78 \pm 2.32$ \\
        H4 & $79.17 \pm 2.80$ & $81.21 \pm 2.25$ & $80.42 \pm 2.30$ & $82.02 \pm 2.13$ \\
        \bottomrule
    \end{tabular}%
    }
\end{table}

Although the standard practice in identity-preserving video generation models is to evaluate identity similarity using the same reference images that the model is conditioned on, to assess whether our gains depend on overlap between conditioning and evaluation images, Table~\ref{tab:heldout-expression-models} evaluates all models against four held-out expressions that are not used for conditioning.  The results demonstrate that the model ordering remains consistent with the original evaluation.

\begin{table}[htbp]
    \centering
    \caption{Nearest-gallery identity similarity using four held-out expressions. None of these photographs is used to condition any model. Mild, Intense, and Extreme restrict evaluation to frames in the corresponding expression-intensity regime.}
    \label{tab:heldout-expression-models}
    \resizebox{\textwidth}{!}{%
    \begin{tabular}{lrrrrrrrr}
        \toprule
        & \multicolumn{2}{c}{All frames} & \multicolumn{2}{c}{Mild} & \multicolumn{2}{c}{Intense} & \multicolumn{2}{c}{Extreme} \\
        \cmidrule(lr){2-3} \cmidrule(lr){4-5} \cmidrule(lr){6-7} \cmidrule(l){8-9}
        Model & $s_{\mathrm{Arc}}^* \uparrow$ & $s_{\mathrm{Cur}}^* \uparrow$ & $s_{\mathrm{Arc}}^* \uparrow$ & $s_{\mathrm{Cur}}^* \uparrow$ & $s_{\mathrm{Arc}}^* \uparrow$ & $s_{\mathrm{Cur}}^* \uparrow$ & $s_{\mathrm{Arc}}^* \uparrow$ & $s_{\mathrm{Cur}}^* \uparrow$ \\
        \midrule
        Concat-ID & 54.75 & 53.45 & 65.64 & 64.97 & 54.43 & 52.81 & 42.24 & 41.05 \\
        Phantom & 61.77 & 62.11 & 73.01 & 72.99 & 66.77 & 66.96 & 56.20 & 56.71 \\
        ConsisID & 30.41 & 29.72 & 47.02 & 46.89 & 36.45 & 35.91 & 22.90 & 22.02 \\
        Stand-In & 51.87 & 51.50 & 74.53 & 75.33 & 54.02 & 53.34 & 41.28 & 40.74 \\
        AnyID & 59.24 & 58.32 & 70.21 & 69.37 & 63.09 & 62.13 & 51.25 & 50.39 \\
        Gen-4.5 & 52.26 & 51.72 & 69.09 & 69.69 & 53.82 & 53.29 & 38.63 & 37.27 \\
        \midrule
        Ours (Synthetic) & 70.99 & 71.18 & 70.38 & 69.48 & 72.05 & 72.38 & 66.20 & 66.85 \\
        Ours (Real) & \textbf{76.29} & \textbf{76.37} & \textbf{77.98} & \textbf{77.75} & \textbf{76.20} & \textbf{76.35} & \textbf{70.27} & \textbf{70.92} \\
        \bottomrule
    \end{tabular}
    }
\end{table}

\subsection{Results on OpenS2V-Eval}
OpenS2V-Eval~\citep{yuan2025opens2v} evaluates general-purpose subject-to-video generation. It is therefore not designed specifically for the task studied in this paper: preserving facial identity under controlled, high-intensity expression changes using multiple reference images. In particular, its prompts do not systematically elicit high-intensity expressions, and it provides only one reference image per subject. Still, for context, we evaluate on the full Human-Domain protocol of 60 videos: 30 single-face and 30 single-human entries, each with one reference image and one text prompt. Because the benchmark provides only one reference per entry, we construct an eight-image gallery using our LivePortrait-based synthetic-reference pipeline.

We follow the official OpenS2V-Eval Face Similarity evaluator and leaderboard aggregation. The evaluator uniformly samples 32 frames, computes non-negative CurricularFace cosine similarity to the benchmark reference image for frames with a detected face, and averages the per-frame scores within each video. A video with no detected face receives a score of zero. The reported score is the mean over all 60 videos, multiplied by 100. As shown in Table~\ref{tab:opens2v-facesim-results}, our model obtains a much higher Face Similarity score than the baselines.
\begin{table}[hbp]
    \centering
    \caption{Official OpenS2V-Eval Face Similarity on the 60-video Human-Domain protocol (30 single-face and 30 single-human videos). Baseline scores are the published leaderboard values, and ours uses the same official evaluator and aggregation with references generated by LivePortrait.}
    \label{tab:opens2v-facesim-results}
    \begin{tabular}{lr}
        \toprule
        Model & Face Similarity $\uparrow$ \\
        \midrule
        ConsisID & 43.19 \\
        Concat-ID & 50.05 \\
        Stand-In & 47.99 \\
        Phantom-14B & 55.04 \\
        \midrule
        Ours (Synthetic) & \textbf{62.26} \\
        \bottomrule
    \end{tabular}
\end{table}

\section{Verbal Instructions for Gallery Expressions}
\label{sec:gallery-expression-instructions}

Table~\ref{tab:gallery-expression-instructions} maps the neutral reference
and seven expressions shown in the gallery to the verbal instructions used to
capture them.

\begin{table}[htb]
    \centering
    \small
    \caption{Verbal instructions associated with the neutral reference and expression images in the gallery.}
    \label{tab:gallery-expression-instructions}
    \begin{tabular}{@{}lp{0.56\linewidth}@{}}
        \toprule
        Gallery expression & Verbal instruction \\
        \midrule
        Neutral (0) & Eyes open, teeth together, and lips closed in a neutral resting expression. \\
        Expression 1 & Eyes squeezed shut with raised cheeks, a wrinkled nose, smiling lip corners, and slightly parted lips. \\
        Expression 2 & Eyes directed upward while the lips are pushed forward into a rounded funnel shape. \\
        Expression 3 & Brows drawn together near the center while the mouth stretches open wide. \\
        Expression 4 & Inner brows raised with one or both lip corners tightened backward into a restrained, dimpling expression. \\
        Expression 5 & Eyes looking up and to the left while the jaw shifts toward the right. \\
        Expression 6 & Upper lip lifted, lower lip pulled downward, and the outer brows raised. \\
        Expression 7 & Upper lip raised, lower lip pulled down, and the mouth stretched very wide open. \\
        \bottomrule
    \end{tabular}
\end{table}

\section{Synthetic Gallery Example}
\label{sec:liveportrait-gallery}
\begin{figure}[htb]
    \centering
    \begin{tabular}{@{}*{7}{c@{\hspace{0.005\linewidth}}}c@{}}
        \includegraphics[width=0.12\linewidth]{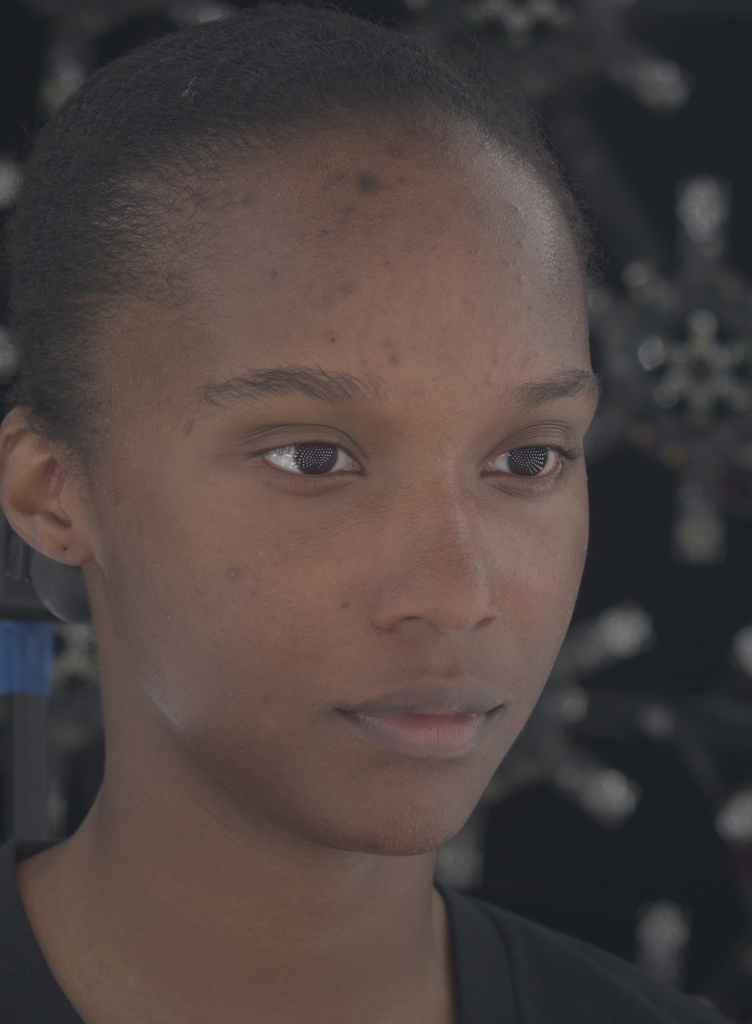} &
        \includegraphics[width=0.12\linewidth]{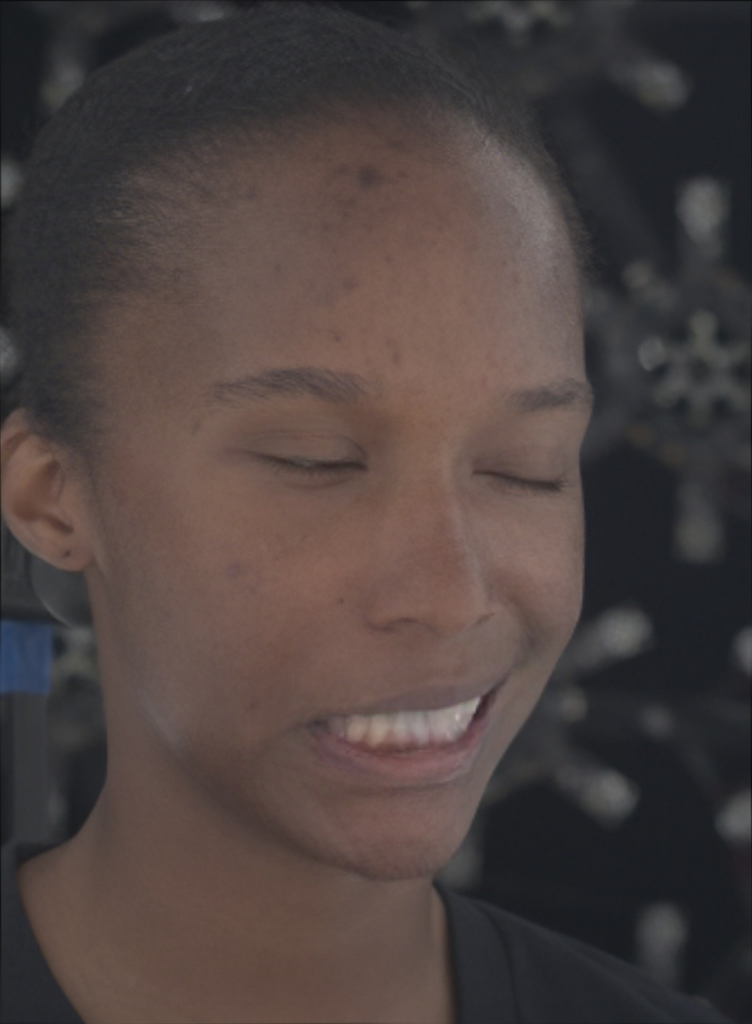} &
        \includegraphics[width=0.12\linewidth]{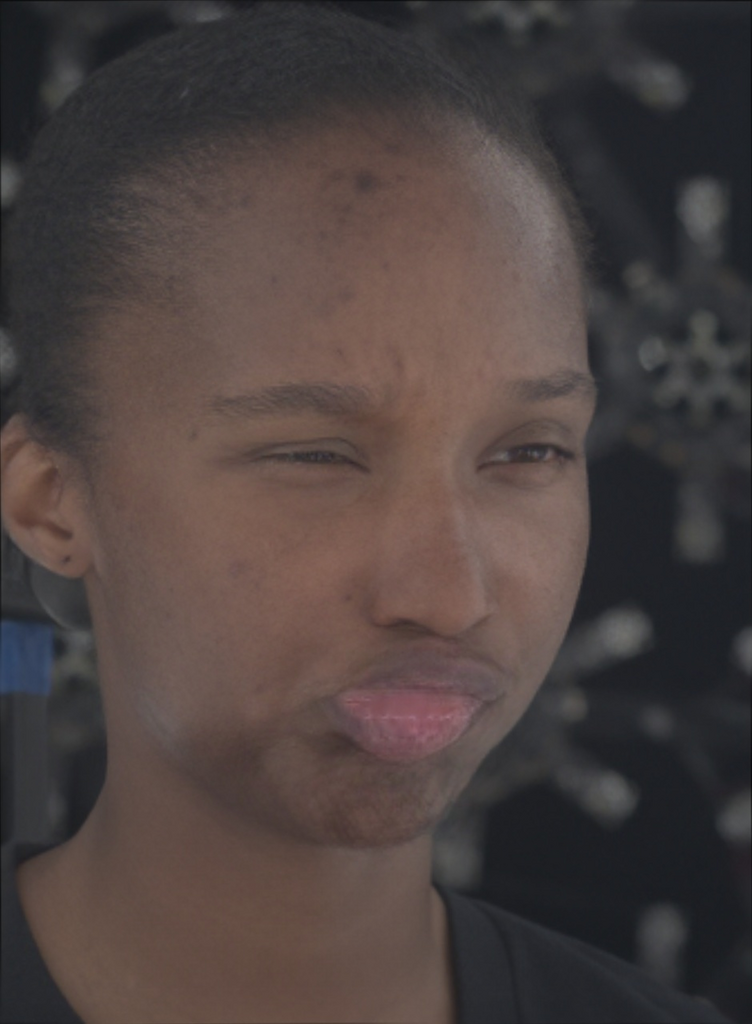} &
        \includegraphics[width=0.12\linewidth]{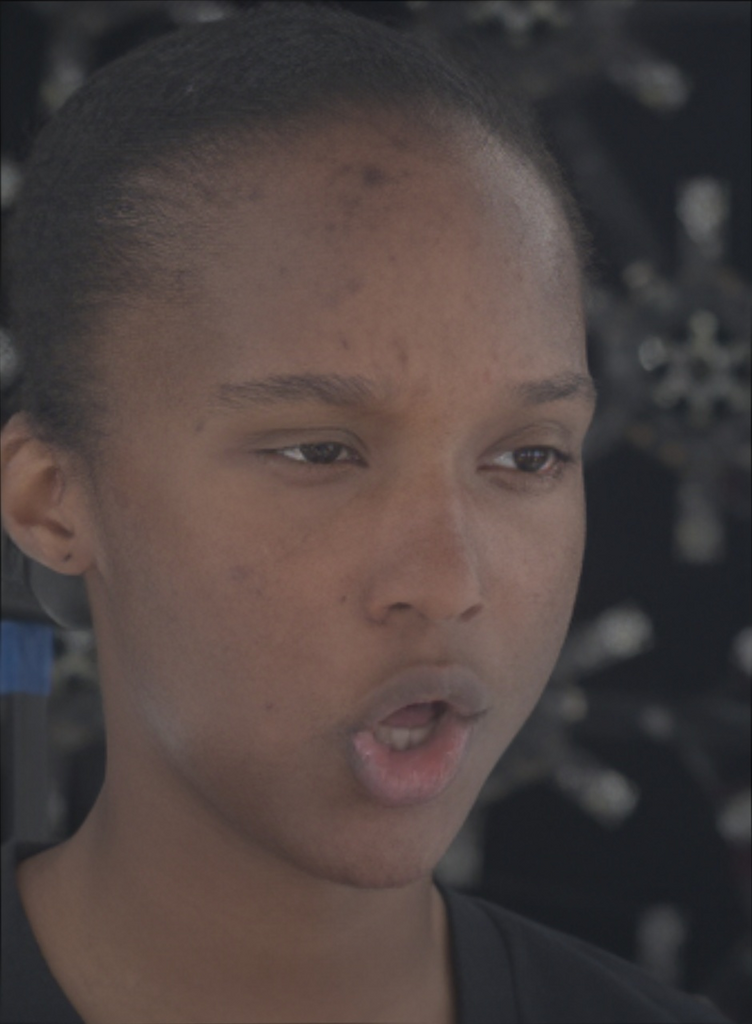} &
        \includegraphics[width=0.12\linewidth]{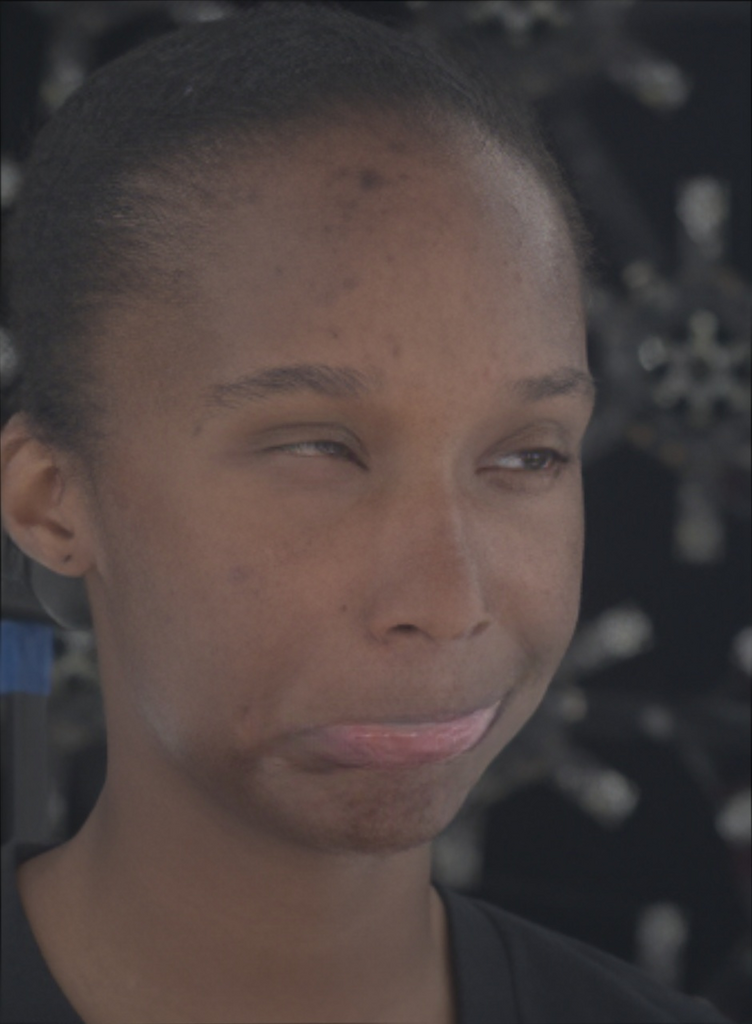} &
        \includegraphics[width=0.12\linewidth]{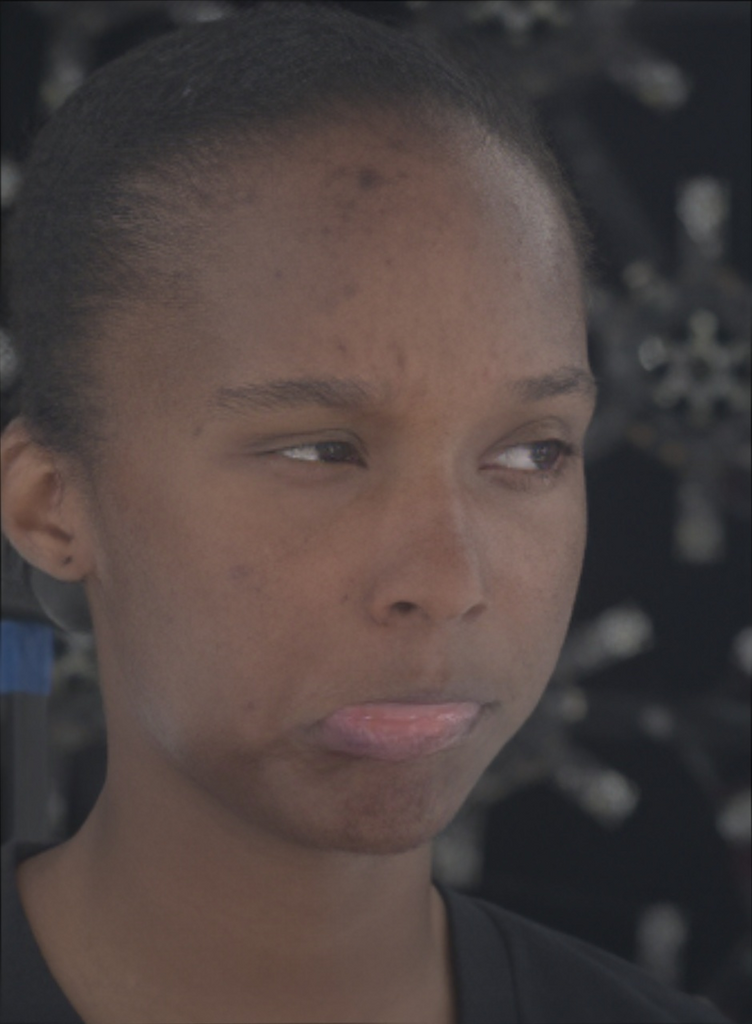} &
        \includegraphics[width=0.12\linewidth]{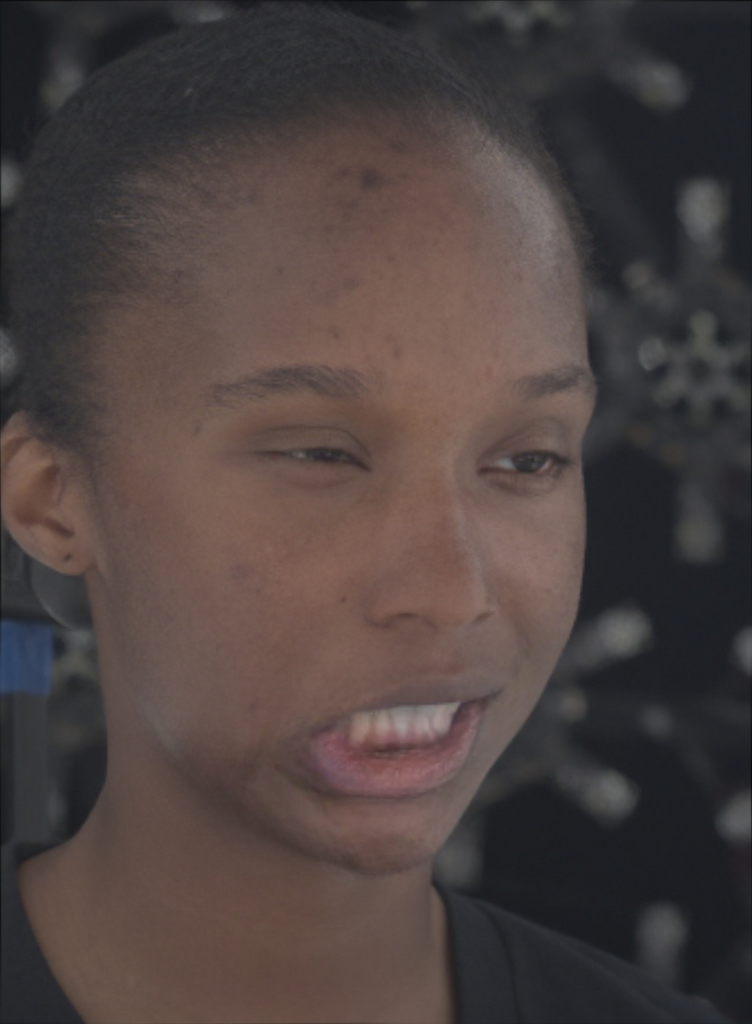} &
        \includegraphics[width=0.12\linewidth]{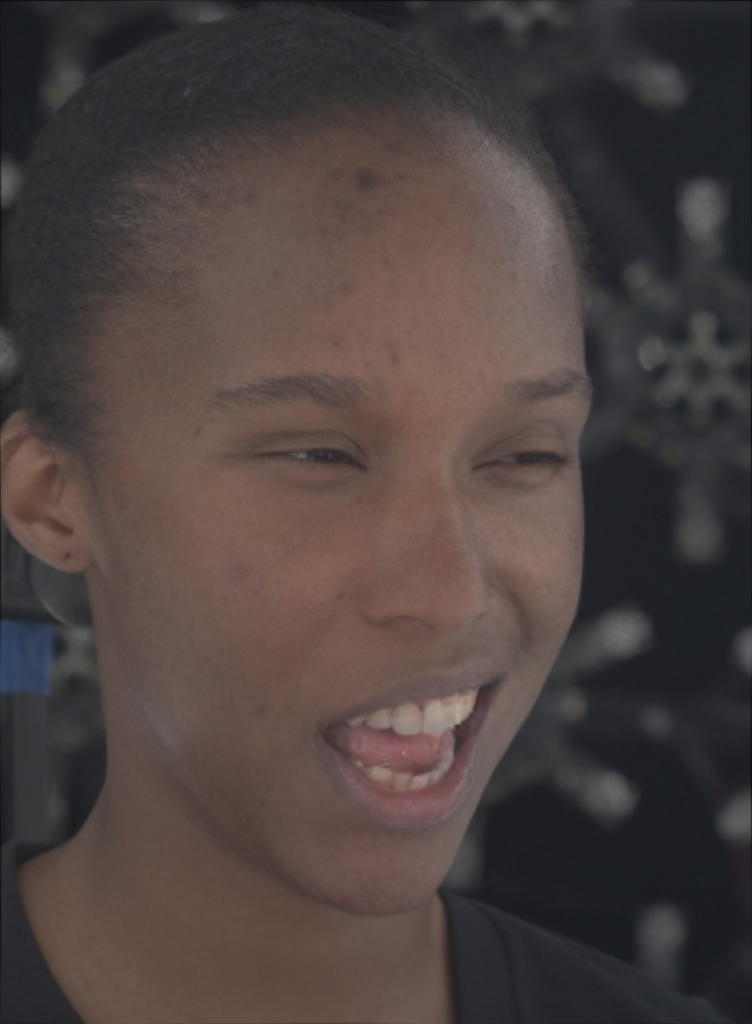} \\
        \scriptsize Neutral 0 &
        \scriptsize Synthetic 1 &
        \scriptsize Synthetic 2 &
        \scriptsize Synthetic 3 &
        \scriptsize Synthetic 4 &
        \scriptsize Synthetic 5 &
        \scriptsize Synthetic 6 &
        \scriptsize Synthetic 7
    \end{tabular}
    \caption{Example of a synthetic expression gallery generated from a single neutral portrait. Although less natural than the real captures in Figure~\ref{fig:lightstage-gallery}, the synthetic gallery provides additional expression information for identity-preserving video generation with the image-domain prior of the facial reenactment model.}
    \label{fig:liveportrait-gallery}
\end{figure}

We show an example of the synthetic gallery generated by LivePortrait in Figure~\ref{fig:liveportrait-gallery}.

\section{Prompt Specification}
\label{sec:eval-prompts}
\subsection{Details of Prompt Generation Process}
We design prompts depicting controlled settings with a fixed camera to evaluate identity preservation under strong expressions without introducing factors such as camera movement or occlusion. We design 4 environments (Table~\ref{tab:envs}) and 2 triggers, and use the 6 basic emotions proposed by \citet{Ekman1992Facial} (see the emotion cues in Table~\ref{tab:emotions}) to compile 48 prompts for each subject. We structure the prompts with the following template:
\begin{quote}
    \texttt{Medium close-up in \{Environment Phrase\}. A [GENDER] \{Environment anchor\}, face relaxed and neutral, looking just past the camera. \{trigger\}, and their face \{emotion verb\}: \{cues\}. The expression builds fast and holds at full intensity; the camera stays still.}
\end{quote}

\begin{table}[htb]
    \centering
    \caption{Environment Phrases}
    \label{tab:envs}
    \begin{tabular}{c|ll}
        \toprule
        Environment & Phrase & Anchor \\
        \midrule
        kitchen & a warmly lit, everyday kitchen & stands at the counter \\
        home office & a quiet, softly lit home office & sits at the desk \\
        living room & a dimly lit, quiet living room & sits on the sofa \\
        cafe & a quiet, softly daylit cafe & sits at a table by the window \\
        \bottomrule
    \end{tabular}
\end{table}

\begin{table}[htb]
    \centering
    \caption{Cues for each emotion}
    \label{tab:emotions}
    \resizebox{\textwidth}{!}{\begin{tabular}{c|l}
        \toprule
        Emotion & Cues \\
        \midrule
        joy & eyes narrow into creases, cheeks lift high, the mouth opens wide showing teeth \\
        surprise & brows arch high, eyes go wide with the whites showing, the jaw drops open \\
        fear & brows raise and pull together, eyes stretch wide, the lips pull back flat and tense as the mouth opens \\
        anger & brows pull down hard, eyes glare, lips press tight and then part to bare clenched teeth, nostrils flare \\
        disgust & nose wrinkles, upper lip lifts, lip corners pull down, eyes squint \\
        sadness & inner brows lift and knit together, lip corners pull down, the chin tightens and trembles, eyes fill with tears \\
        \bottomrule
    \end{tabular}
    }
\end{table}
\FloatBarrier
\subsection{Full List of Prompts}
\begin{enumerate}
    \item Medium close-up in a warmly lit, everyday kitchen. A [GENDER] stands at the counter, face relaxed and neutral, looking just past the camera. Someone just out of frame tells a joke, and their face breaks into uncontrollable laughter: eyes narrow into creases, cheeks lift high, the mouth opens wide showing teeth. The expression builds fast and holds at full intensity; the camera stays still.
    \item Medium close-up in a warmly lit, everyday kitchen. A [GENDER] stands at the counter, face relaxed and neutral, looking just past the camera. They read a message on their phone that is absurdly funny, and their face breaks into uncontrollable laughter: eyes narrow into creases, cheeks lift high, the mouth opens wide showing teeth. The expression builds fast and holds at full intensity; the camera stays still.
    \item Medium close-up in a quiet, softly lit home office. A [GENDER] sits at the desk, face relaxed and neutral, looking just past the camera. Someone just out of frame tells a joke, and their face breaks into uncontrollable laughter: eyes narrow into creases, cheeks lift high, the mouth opens wide showing teeth. The expression builds fast and holds at full intensity; the camera stays still.
    \item Medium close-up in a quiet, softly lit home office. A [GENDER] sits at the desk, face relaxed and neutral, looking just past the camera. They read a message on their phone that is absurdly funny, and their face breaks into uncontrollable laughter: eyes narrow into creases, cheeks lift high, the mouth opens wide showing teeth. The expression builds fast and holds at full intensity; the camera stays still.
    \item Medium close-up in a dimly lit, quiet living room. A [GENDER] sits on the sofa, face relaxed and neutral, looking just past the camera. Someone just out of frame tells a joke, and their face breaks into uncontrollable laughter: eyes narrow into creases, cheeks lift high, the mouth opens wide showing teeth. The expression builds fast and holds at full intensity; the camera stays still.
    \item Medium close-up in a dimly lit, quiet living room. A [GENDER] sits on the sofa, face relaxed and neutral, looking just past the camera. They read a message on their phone that is absurdly funny, and their face breaks into uncontrollable laughter: eyes narrow into creases, cheeks lift high, the mouth opens wide showing teeth. The expression builds fast and holds at full intensity; the camera stays still.
    \item Medium close-up in a quiet, softly daylit cafe. A [GENDER] sits at a table by the window, face relaxed and neutral, looking just past the camera. Someone just out of frame tells a joke, and their face breaks into uncontrollable laughter: eyes narrow into creases, cheeks lift high, the mouth opens wide showing teeth. The expression builds fast and holds at full intensity; the camera stays still.
    \item Medium close-up in a quiet, softly daylit cafe. A [GENDER] sits at a table by the window, face relaxed and neutral, looking just past the camera. They read a message on their phone that is absurdly funny, and their face breaks into uncontrollable laughter: eyes narrow into creases, cheeks lift high, the mouth opens wide showing teeth. The expression builds fast and holds at full intensity; the camera stays still.
    \item Medium close-up in a warmly lit, everyday kitchen. A [GENDER] stands at the counter, face relaxed and neutral, looking just past the camera. A voice just out of frame blurts out unexpected news, and their face snaps into open-mouthed astonishment: brows arch high, eyes go wide with the whites showing, the jaw drops open. The expression builds fast and holds at full intensity; the camera stays still.
    \item Medium close-up in a warmly lit, everyday kitchen. A [GENDER] stands at the counter, face relaxed and neutral, looking just past the camera. They glance at their phone and see something they never expected, and their face snaps into open-mouthed astonishment: brows arch high, eyes go wide with the whites showing, the jaw drops open. The expression builds fast and holds at full intensity; the camera stays still.
    \item Medium close-up in a quiet, softly lit home office. A [GENDER] sits at the desk, face relaxed and neutral, looking just past the camera. A voice just out of frame blurts out unexpected news, and their face snaps into open-mouthed astonishment: brows arch high, eyes go wide with the whites showing, the jaw drops open. The expression builds fast and holds at full intensity; the camera stays still.
    \item Medium close-up in a quiet, softly lit home office. A [GENDER] sits at the desk, face relaxed and neutral, looking just past the camera. They glance at their phone and see something they never expected, and their face snaps into open-mouthed astonishment: brows arch high, eyes go wide with the whites showing, the jaw drops open. The expression builds fast and holds at full intensity; the camera stays still.
    \item Medium close-up in a dimly lit, quiet living room. A [GENDER] sits on the sofa, face relaxed and neutral, looking just past the camera. A voice just out of frame blurts out unexpected news, and their face snaps into open-mouthed astonishment: brows arch high, eyes go wide with the whites showing, the jaw drops open. The expression builds fast and holds at full intensity; the camera stays still.
    \item Medium close-up in a dimly lit, quiet living room. A [GENDER] sits on the sofa, face relaxed and neutral, looking just past the camera. They glance at their phone and see something they never expected, and their face snaps into open-mouthed astonishment: brows arch high, eyes go wide with the whites showing, the jaw drops open. The expression builds fast and holds at full intensity; the camera stays still.
    \item Medium close-up in a quiet, softly daylit cafe. A [GENDER] sits at a table by the window, face relaxed and neutral, looking just past the camera. A voice just out of frame blurts out unexpected news, and their face snaps into open-mouthed astonishment: brows arch high, eyes go wide with the whites showing, the jaw drops open. The expression builds fast and holds at full intensity; the camera stays still.
    \item Medium close-up in a quiet, softly daylit cafe. A [GENDER] sits at a table by the window, face relaxed and neutral, looking just past the camera. They glance at their phone and see something they never expected, and their face snaps into open-mouthed astonishment: brows arch high, eyes go wide with the whites showing, the jaw drops open. The expression builds fast and holds at full intensity; the camera stays still.
    \item Medium close-up in a warmly lit, everyday kitchen. A [GENDER] stands at the counter, face relaxed and neutral, looking just past the camera. Something crashes loudly just out of frame, and their face freezes in wide-eyed fear: brows raise and pull together, eyes stretch wide, the lips pull back flat and tense as the mouth opens. The expression builds fast and holds at full intensity; the camera stays still.
    \item Medium close-up in a warmly lit, everyday kitchen. A [GENDER] stands at the counter, face relaxed and neutral, looking just past the camera. Their phone lights up with a message that terrifies them, and their face freezes in wide-eyed fear: brows raise and pull together, eyes stretch wide, the lips pull back flat and tense as the mouth opens. The expression builds fast and holds at full intensity; the camera stays still.
    \item Medium close-up in a quiet, softly lit home office. A [GENDER] sits at the desk, face relaxed and neutral, looking just past the camera. Something crashes loudly just out of frame, and their face freezes in wide-eyed fear: brows raise and pull together, eyes stretch wide, the lips pull back flat and tense as the mouth opens. The expression builds fast and holds at full intensity; the camera stays still.
    \item Medium close-up in a quiet, softly lit home office. A [GENDER] sits at the desk, face relaxed and neutral, looking just past the camera. Their phone lights up with a message that terrifies them, and their face freezes in wide-eyed fear: brows raise and pull together, eyes stretch wide, the lips pull back flat and tense as the mouth opens. The expression builds fast and holds at full intensity; the camera stays still.
    \item Medium close-up in a dimly lit, quiet living room. A [GENDER] sits on the sofa, face relaxed and neutral, looking just past the camera. Something crashes loudly just out of frame, and their face freezes in wide-eyed fear: brows raise and pull together, eyes stretch wide, the lips pull back flat and tense as the mouth opens. The expression builds fast and holds at full intensity; the camera stays still.
    \item Medium close-up in a dimly lit, quiet living room. A [GENDER] sits on the sofa, face relaxed and neutral, looking just past the camera. Their phone lights up with a message that terrifies them, and their face freezes in wide-eyed fear: brows raise and pull together, eyes stretch wide, the lips pull back flat and tense as the mouth opens. The expression builds fast and holds at full intensity; the camera stays still.
    \item Medium close-up in a quiet, softly daylit cafe. A [GENDER] sits at a table by the window, face relaxed and neutral, looking just past the camera. Something crashes loudly just out of frame, and their face freezes in wide-eyed fear: brows raise and pull together, eyes stretch wide, the lips pull back flat and tense as the mouth opens. The expression builds fast and holds at full intensity; the camera stays still.
    \item Medium close-up in a quiet, softly daylit cafe. A [GENDER] sits at a table by the window, face relaxed and neutral, looking just past the camera. Their phone lights up with a message that terrifies them, and their face freezes in wide-eyed fear: brows raise and pull together, eyes stretch wide, the lips pull back flat and tense as the mouth opens. The expression builds fast and holds at full intensity; the camera stays still.
    \item Medium close-up in a warmly lit, everyday kitchen. A [GENDER] stands at the counter, face relaxed and neutral, looking just past the camera. A voice just out of frame says something insulting, and their face hardens into open anger: brows pull down hard, eyes glare, lips press tight and then part to bare clenched teeth, nostrils flare. The expression builds fast and holds at full intensity; the camera stays still.
    \item Medium close-up in a warmly lit, everyday kitchen. A [GENDER] stands at the counter, face relaxed and neutral, looking just past the camera. They read a message on their phone that makes their blood boil, and their face hardens into open anger: brows pull down hard, eyes glare, lips press tight and then part to bare clenched teeth, nostrils flare. The expression builds fast and holds at full intensity; the camera stays still.
    \item Medium close-up in a quiet, softly lit home office. A [GENDER] sits at the desk, face relaxed and neutral, looking just past the camera. A voice just out of frame says something insulting, and their face hardens into open anger: brows pull down hard, eyes glare, lips press tight and then part to bare clenched teeth, nostrils flare. The expression builds fast and holds at full intensity; the camera stays still.
    \item Medium close-up in a quiet, softly lit home office. A [GENDER] sits at the desk, face relaxed and neutral, looking just past the camera. They read a message on their phone that makes their blood boil, and their face hardens into open anger: brows pull down hard, eyes glare, lips press tight and then part to bare clenched teeth, nostrils flare. The expression builds fast and holds at full intensity; the camera stays still.
    \item Medium close-up in a dimly lit, quiet living room. A [GENDER] sits on the sofa, face relaxed and neutral, looking just past the camera. A voice just out of frame says something insulting, and their face hardens into open anger: brows pull down hard, eyes glare, lips press tight and then part to bare clenched teeth, nostrils flare. The expression builds fast and holds at full intensity; the camera stays still.
    \item Medium close-up in a dimly lit, quiet living room. A [GENDER] sits on the sofa, face relaxed and neutral, looking just past the camera. They read a message on their phone that makes their blood boil, and their face hardens into open anger: brows pull down hard, eyes glare, lips press tight and then part to bare clenched teeth, nostrils flare. The expression builds fast and holds at full intensity; the camera stays still.
    \item Medium close-up in a quiet, softly daylit cafe. A [GENDER] sits at a table by the window, face relaxed and neutral, looking just past the camera. A voice just out of frame says something insulting, and their face hardens into open anger: brows pull down hard, eyes glare, lips press tight and then part to bare clenched teeth, nostrils flare. The expression builds fast and holds at full intensity; the camera stays still.
    \item Medium close-up in a quiet, softly daylit cafe. A [GENDER] sits at a table by the window, face relaxed and neutral, looking just past the camera. They read a message on their phone that makes their blood boil, and their face hardens into open anger: brows pull down hard, eyes glare, lips press tight and then part to bare clenched teeth, nostrils flare. The expression builds fast and holds at full intensity; the camera stays still.
    \item Medium close-up in a warmly lit, everyday kitchen. A [GENDER] stands at the counter, face relaxed and neutral, looking just past the camera. A foul smell drifts in from just out of frame, and their face twists into intense disgust: nose wrinkles, upper lip lifts, lip corners pull down, eyes squint. The expression builds fast and holds at full intensity; the camera stays still.
    \item Medium close-up in a warmly lit, everyday kitchen. A [GENDER] stands at the counter, face relaxed and neutral, looking just past the camera. They look at their phone and see something revolting, and their face twists into intense disgust: nose wrinkles, upper lip lifts, lip corners pull down, eyes squint. The expression builds fast and holds at full intensity; the camera stays still.
    \item Medium close-up in a quiet, softly lit home office. A [GENDER] sits at the desk, face relaxed and neutral, looking just past the camera. A foul smell drifts in from just out of frame, and their face twists into intense disgust: nose wrinkles, upper lip lifts, lip corners pull down, eyes squint. The expression builds fast and holds at full intensity; the camera stays still.
    \item Medium close-up in a quiet, softly lit home office. A [GENDER] sits at the desk, face relaxed and neutral, looking just past the camera. They look at their phone and see something revolting, and their face twists into intense disgust: nose wrinkles, upper lip lifts, lip corners pull down, eyes squint. The expression builds fast and holds at full intensity; the camera stays still.
    \item Medium close-up in a dimly lit, quiet living room. A [GENDER] sits on the sofa, face relaxed and neutral, looking just past the camera. A foul smell drifts in from just out of frame, and their face twists into intense disgust: nose wrinkles, upper lip lifts, lip corners pull down, eyes squint. The expression builds fast and holds at full intensity; the camera stays still.
    \item Medium close-up in a dimly lit, quiet living room. A [GENDER] sits on the sofa, face relaxed and neutral, looking just past the camera. They look at their phone and see something revolting, and their face twists into intense disgust: nose wrinkles, upper lip lifts, lip corners pull down, eyes squint. The expression builds fast and holds at full intensity; the camera stays still.
    \item Medium close-up in a quiet, softly daylit cafe. A [GENDER] sits at a table by the window, face relaxed and neutral, looking just past the camera. A foul smell drifts in from just out of frame, and their face twists into intense disgust: nose wrinkles, upper lip lifts, lip corners pull down, eyes squint. The expression builds fast and holds at full intensity; the camera stays still.
    \item Medium close-up in a quiet, softly daylit cafe. A [GENDER] sits at a table by the window, face relaxed and neutral, looking just past the camera. They look at their phone and see something revolting, and their face twists into intense disgust: nose wrinkles, upper lip lifts, lip corners pull down, eyes squint. The expression builds fast and holds at full intensity; the camera stays still.
    \item Medium close-up in a warmly lit, everyday kitchen. A [GENDER] stands at the counter, face relaxed and neutral, looking just past the camera. A voice just out of frame delivers devastating news, and their face crumples into open grief: inner brows lift and knit together, lip corners pull down, the chin tightens and trembles, eyes fill with tears. The expression builds fast and holds at full intensity; the camera stays still.
    \item Medium close-up in a warmly lit, everyday kitchen. A [GENDER] stands at the counter, face relaxed and neutral, looking just past the camera. They read a message on their phone with devastating news, and their face crumples into open grief: inner brows lift and knit together, lip corners pull down, the chin tightens and trembles, eyes fill with tears. The expression builds fast and holds at full intensity; the camera stays still.
    \item Medium close-up in a quiet, softly lit home office. A [GENDER] sits at the desk, face relaxed and neutral, looking just past the camera. A voice just out of frame delivers devastating news, and their face crumples into open grief: inner brows lift and knit together, lip corners pull down, the chin tightens and trembles, eyes fill with tears. The expression builds fast and holds at full intensity; the camera stays still.
    \item Medium close-up in a quiet, softly lit home office. A [GENDER] sits at the desk, face relaxed and neutral, looking just past the camera. They read a message on their phone with devastating news, and their face crumples into open grief: inner brows lift and knit together, lip corners pull down, the chin tightens and trembles, eyes fill with tears. The expression builds fast and holds at full intensity; the camera stays still.
    \item Medium close-up in a dimly lit, quiet living room. A [GENDER] sits on the sofa, face relaxed and neutral, looking just past the camera. A voice just out of frame delivers devastating news, and their face crumples into open grief: inner brows lift and knit together, lip corners pull down, the chin tightens and trembles, eyes fill with tears. The expression builds fast and holds at full intensity; the camera stays still.
    \item Medium close-up in a dimly lit, quiet living room. A [GENDER] sits on the sofa, face relaxed and neutral, looking just past the camera. They read a message on their phone with devastating news, and their face crumples into open grief: inner brows lift and knit together, lip corners pull down, the chin tightens and trembles, eyes fill with tears. The expression builds fast and holds at full intensity; the camera stays still.
    \item Medium close-up in a quiet, softly daylit cafe. A [GENDER] sits at a table by the window, face relaxed and neutral, looking just past the camera. A voice just out of frame delivers devastating news, and their face crumples into open grief: inner brows lift and knit together, lip corners pull down, the chin tightens and trembles, eyes fill with tears. The expression builds fast and holds at full intensity; the camera stays still.
    \item Medium close-up in a quiet, softly daylit cafe. A [GENDER] sits at a table by the window, face relaxed and neutral, looking just past the camera. They read a message on their phone with devastating news, and their face crumples into open grief: inner brows lift and knit together, lip corners pull down, the chin tightens and trembles, eyes fill with tears. The expression builds fast and holds at full intensity; the camera stays still.

\end{enumerate}

\end{document}